\documentclass[lettersize,journal]{IEEEtran}
\usepackage{amsmath,amsfonts}
\usepackage{algorithmic}
\usepackage{array}
\usepackage[caption=false,font=normalsize,labelfont=sf,textfont=sf]{subfig}
\usepackage[caption=false,font=normalsize,labelfont=sf,textfont=sf]{subfig}
\usepackage{textcomp}
\usepackage{stfloats}
\usepackage{url}
\usepackage{verbatim}
\usepackage{graphicx}
\usepackage{cite}
\usepackage{ragged2e} 
\usepackage{booktabs,makecell, multirow, tabularx}
\usepackage{hyperref}
\usepackage{xcolor}
\usepackage[ruled,linesnumbered]{algorithm2e}
\begin{document}
	
	\title{FDCE-Net: Underwater Image Enhancement with Embedding Frequency and Dual Color Encoder}
	\author{\IEEEauthorblockN{
			Zheng Cheng, Guodong Fan, Jingchun Zhou, Min Gan, ~\IEEEmembership{Senior Member,~IEEE}, \\and C. L. Philip Chen, ~\IEEEmembership{Fellow,~IEEE}}
		\thanks{This work was supported in part by the National
			Natural Science Foundation of China under Grant 62073082, Grant 71701136,
			Grant U1813203, and Grant U1801262; in part by the Taishan Scholar
			Program of Shandong Province; in part by the Natural Science Foundation of
			Guangdong Province under Grant 2022A1515011009; in part by the Science
			and Technology Major Project of Guangzhou under Grant 202007030006;
			in part by the Science and Technology Development Fund, Macau, under
			Grant 0119/2018/A3; and in part by the Guangdong Basic and Applied Basic
			Research Foundation under Grant 2021A1515011860. \textit{(Zheng Cheng and Guodong Fan contributed equally to this work.)(Corresponding author: Guodong Fan.)}
		}
		\thanks{Zheng Cheng, Guodong Fan and Min Gan are with the College of Computer Science \& Technology, Qingdao University, Qingdao 266071, China(e-mail:2022020691@qdu.edu.cn; fgd96@outlook.com; aganmin@aliyun.com).}
		
		\thanks{Jingchun Zhou is with College of Information Science and Technology, Dalian Maritime University, Dalian 116026, China(e-mail:zhoujingchun03@qq.com)}
		
		\thanks{C. L. Philip Chen is with the College of Computer Science and Technology, Qingdao University, Qingdao 266071, China, and also with the School of Computer Science and Engineering, South China University of Technology, Guangzhou 510641, China (e-mail: philip.chen@ieee.org).}

	}

\markboth{Journal of \LaTeX\ Class Files,~Vol.~14, No.~8, August~2021}%
{Shell \MakeLowercase{\textit{et al.}}: A Sample Article Using IEEEtran.cls for IEEE Journals}

\IEEEpubid{}

\maketitle

\begin{abstract}
	
	Underwater images often suffer from various issues such as low brightness, color shift, blurred details, and noise due to light absorption and scattering caused by water and suspended particles. 
	Previous underwater image enhancement (UIE) methods have primarily focused on spatial domain enhancement, neglecting the frequency domain information inherent in the images.
	{However, the degradation factors of underwater images are closely intertwined in the spatial domain.}
	Although certain methods focus on enhancing images in the frequency domain, they overlook the inherent relationship between the image degradation factors and the information present in the frequency domain. {As a result, these methods frequently enhance certain attributes of the improved image while inadequately addressing or even exacerbating other attributes.}
	Moreover, many existing methods heavily rely on prior knowledge to address color shift problems in underwater images, limiting their flexibility and robustness. In order to overcome these limitations, we propose the Embedding Frequency and Dual Color Encoder Network (FDCE-Net) in our paper.
	The FDCE-Net consists of two main structures:
	(1) Frequency Spatial Network (FS-Net) aims to achieve initial enhancement by utilizing our designed Frequency Spatial Residual Block (FSRB) to decouple image degradation factors in the frequency domain and enhance different attributes separately.
	(2) To tackle the color shift issue, we introduce the Dual-Color Encoder (DCE). The DCE establishes correlations between color and semantic representations through cross-attention and leverages multi-scale image features to guide the optimization of adaptive color query.
	The final enhanced images are generated by combining the outputs of FS-Net and DCE through a fusion network. These images exhibit rich details, clear textures, low noise and natural colors.
	Extensive experiments demonstrate that our FDCE-Net outperforms state-of-the-art (SOTA) methods in terms of both visual quality and quantitative metrics. The code of our model is publicly available at: \url{https://github.com/Alexande-rChan/FDCE-Net}.
\end{abstract}

\begin{IEEEkeywords}
	Underwater Image Enhancement, Frequency Convolution, Transformer
\end{IEEEkeywords}

\section{Introduction}
\IEEEPARstart{T}{he} oceans cover approximately 70\% of the Earth's surface, playing a crucial role in sustaining life and offering vast resources. As the exploration of the oceans gains increasing importance worldwide, there is a growing emphasis on visually-guided Autonomous Underwater Vehicles (AUVs) and Remotely Operated Vehicles (ROVs) for various tasks such as submarine cable maintenance, inspection of natural gas pipelines, shipwreck discovery, archaeology, and seabed mapping. However, underwater imaging faces significant challenges due to the complex underwater environment and lighting conditions, resulting in degradation characterized by fuzzy details, color shifts, and low brightness and high noise. Despite the use of advanced imaging systems, extracting valuable information from degraded underwater images remains highly challenging.

The absorption of light by water varies with wavelength. As the water depth increases, red light is absorbed first due to its longer wavelength. Green and blue light, with shorter wavelengths, can penetrate the water to a greater extent, resulting in a blue-green tint in underwater images. Additionally, the scattering effect of suspended particles in the water causes scene blurring, making it challenging to apply image enhancement methods used in other tasks, such as low-light image enhancement, image dehazing or image denoising, to UIE task.

In the early days, traditional methods dominated the task of UIE. These methods relied on physical models and involved establishing degradation models such as the Jaffe-McGlamery underwater image formation model\cite{jaffe1990computer} and atmospheric scattering models initially proposed by McCartney\cite{mccartney1976optics}, and later refined by Narasimhan\cite{narasimhan2003contrast} and Nayar\cite{nayar1999vision}. By computing parameters and solving the inverse problem of image formation, these methods could obtain enhanced images. However, model-based approaches relied on prior knowledge and assumptions about environmental conditions, and the optimal choice of priors remained unclear. Additionally, the degree to which these priors conformed to image statistics and their impact on image enhancement performance remained unknown, while the process of parameterizing the models introduced computational complexity.

Traditional non-model-based methods focus on manipulating the pixel values of an image to enhance its visualization. These methods include techniques like histogram equalization\cite{hummel1975image,pizer1987adaptive} and color correction\cite{iqbal2010enhancing}. However, these methods have limitations including sensitivity to specific conditions, information loss, color distortions, and difficulty in recovering lost details.
In recent years, deep learning-based methods for image enhancement and restoration have shown remarkable performance in various domains, including image super-resolution\cite{9992208,10355918,10380717,10314558,10382425}, low-light image enhancement\cite{fan2022multiscale,ma2022low,ma2023bilevel,10398262} and video processing\cite{10462144,10336558,10145763}. These advancements have also been extended to the task of UIE\cite{liu2022twin,jiang2022target,jiang2022underwater,10087020,9547730,10382428,ZHANG2024685}. Some approaches combine deep learning with physical models, utilizing the powerful feature extraction capabilities of convolutional neural networks (CNNs) to estimate parameter values, such as the transmission map, within the imaging model. Others directly generate the enhanced image using CNNs' strong fitting abilities. With the utilization of large-scale datasets and novel network architectures, these methods have surpassed previous techniques in terms of performance and robustness.

\begin{figure}[t]
	\centering
	\includegraphics[width=0.42\textwidth]{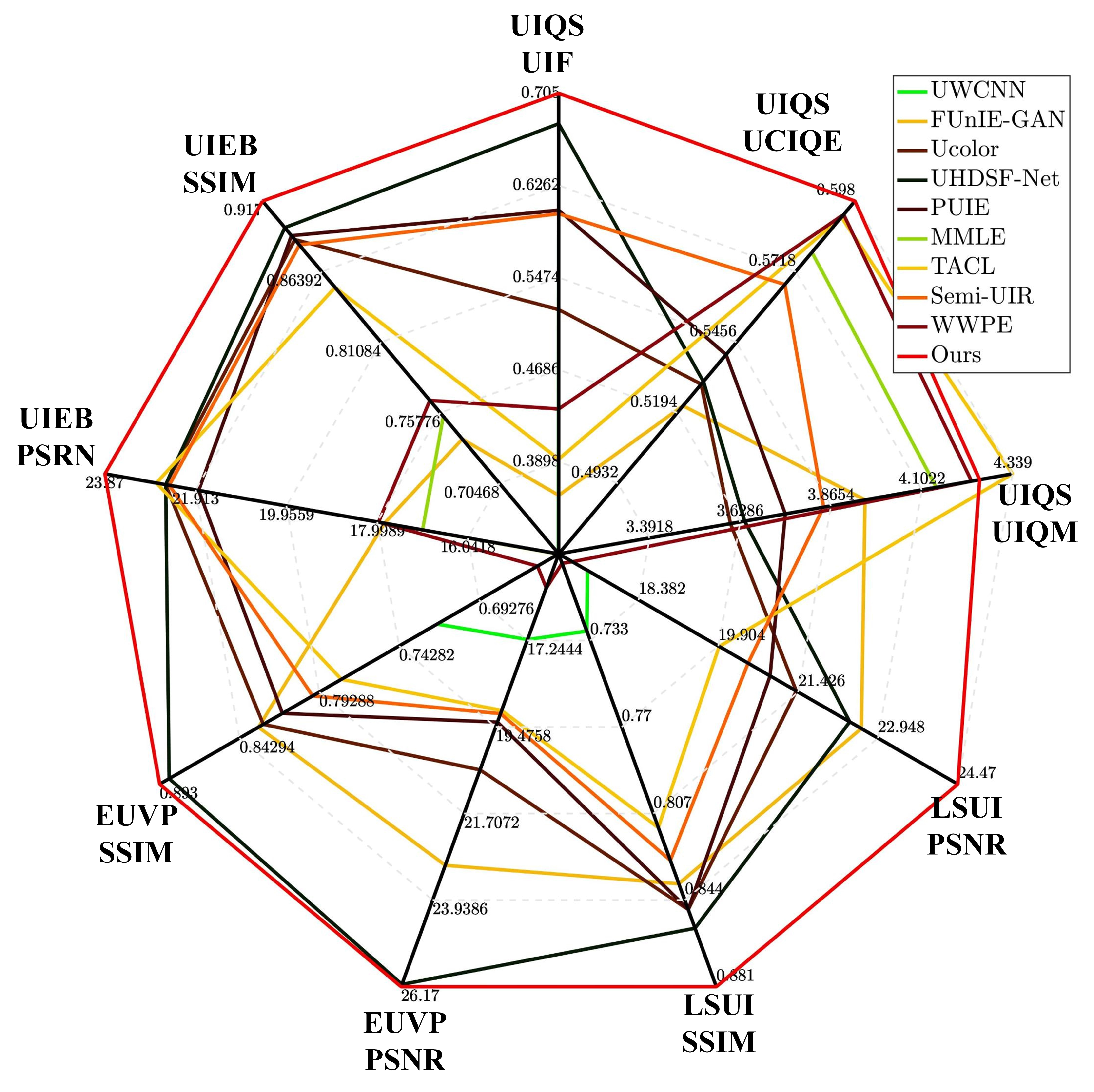}
	\caption{We selected the PSNR and SSIM metrics for the paired dataset and the UIQM, UCIQE and UIF metrics for the unpaired dataset to plot the radar charts, with the coordinate points farther away from the center representing the better performance in that particular metric.}
	\label{fig:Radar}
\end{figure}

\begin{figure*}[b]
	\centering
	\includegraphics[width=1\textwidth]{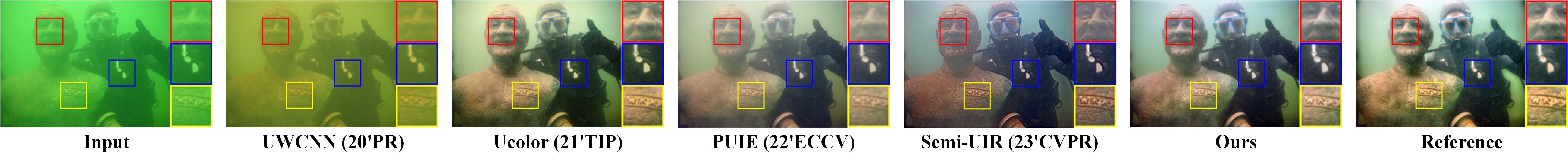}
	\caption{Visual results of SOTA UIE methods. We can see that existing methods cannot cope with challenging underwater image well, where color, brightness, noise and texture details are not handled in a balanced manner.}
	\label{fig:Imbalance}
\end{figure*}
Despite the progress made in deep-learning-based UIE methods, there are still several challenges that need to be addressed. {The core challenge of existing UIE methods is the issue of uneven image enhancement, as mentioned in our introduction. Current UIE methods are unable to address the multitude of degradation factors that underwater images face simultaneously. In the process of enhancing underwater images, it's difficult to avoid compromising one aspect for another, such as increasing brightness at the expense of introducing more noise, or achieving better color restoration while failing to maintain texture clarity. The reason for these phenomena is that most existing methods enhance images in the spatial domain, where various degradation factors in underwater images are closely interconnected, making it challenging to simultaneously enhance different aspects.
	Consequently, these methods often improve certain attributes of the enhanced image while inadequately addressing or even exacerbating other attributes. For instance, increasing picture brightness could lead to the amplification of noise, and restoring image color would unavoidably result in blurred textures or the introduction of artifacts, as we can see in Fig. \ref{fig:Imbalance}.}
{Moreover, many existing methods heavily rely on prior knowledge for color enhancement, but these priors can not be universally applicable to all underwater scenarios. As a result, these models lack the ability to generalize and adapt to environmental changes and complex scenarios.}

{In this paper, we propose an end-to-end UIE network, FDCE-Net, comprising two main components: FS-Net and DCE. FS-Net decouples the degradation factors present in underwater images and enhances them individually, avoiding the trade-off between different attributes. This approach allows for a more balanced enhancement. To eliminate reliance on prior knowledge for color restoration, we introduce the DCE, which integrates Transformer and CNN architectures to synergistically correlate color with multi-scale semantic representations through cross-attention. Leveraging multi-scale image features to learn color queries, our method can improve color recovery in complex underwater scenes. We employ a three-step approach for the UIE task: initially enhancing images with FS-Net, utilizing DCE to capture the relationship between spatial regions and color queries for better color enhancement, and finally refining overall image details and colors through a fusion network, combining the preliminary enhanced images with insights gained from DCE, resulting in visually satisfactory results. he main contributions of this paper can be summarized as follows:}

\begin{enumerate}
	\item 	{We propose FS-Net, which can decouple the factors causing damage to underwater images in the frequency domain. By employing a divide-and-conquer approach to enhance different aspects of image damage, FS-Net overcomes the common problem of enhancement imbalance in UIE tasks.}
	
	\item We propose DCE, which combines CNN and Transformer to capture the relationship between spatial regions and color queries. DCE learns adaptive color queries from different semantic features, which plays a crucial role in improving the final image color enhancement results.
	
	%
	
	\item {FDCE-Net employs a progressive enhancement strategy, initially enhancing the image coarsely before integrating color features. This ensures color consistency while enhancing structure and texture details. Experimental results demonstrate superior performance compared to existing SOTA methods in both visual quality and quantitative metrics, as illustrated by radar chart analysis (Fig. \ref{fig:Radar}).}
\end{enumerate}

\section{Related Works}
In this section, we provide a brief review of the development of UIE algorithms, as well as the application of Transformer and Fourier Transform in image processing, highlighting their relationship with our proposed method.
\subsection{Underwater Image Enhancement Methods}
{\textbf{Traditional model-based methods} in UIE rely on manually designed priors to estimate unknown parameters of the underwater imaging model\cite{jaffe1990computer, mccartney1976optics}. 
	For example, Berman et al. extended the haze-lines model to consider wavelength-dependent attenuation and oceanographer-classified water types\cite{berman2020underwater}. The method can automatically selects the best parameters based on color distribution.
	Fu et al. proposed a retinex-based UIE method that includes color correction, layer decomposition, and respective enhancement achieved satisfactory results at that time\cite{fu2014retinex}. 
	Liang et al. introduced GUDCP which employs hierarchical backscattered light estimation, robust transmission estimation, and discrete distance guided color correction effectively addressed low contrast and color cast in underwater images. \cite{liang2021gudcp}
	Zhou et al. proposed UGIF-Net effectively enhances underwater images by addressing strong scattering in color channels through a two-color space approach, guided information, and dense feature extraction\cite{zhou2023underwater}.}

{\textbf{Traditional non-model-based methods}, on the other hand, primarily modify the pixel values of underwater images based on specific attributes such as color distortion and reduced contrast\cite{hummel1975image,pizer1987adaptive,iqbal2010enhancing,hitam2013mixture,li2016underwater,10040560,10457069}. These methods do not explicitly model the physical characteristics of underwater imaging. For instance, histogram equalization (HE) transforms the image histogram for a more balanced distribution\cite{hummel1975image}. Adaptive histogram equalization (AHE) enhances local contrast by adapting the histogram equalization to smaller image regions\cite{pizer1987adaptive}. 
	Iqbal et al. presented the Unsupervised Colour Correction Method (UCM) effectively addresses the challenges of high bluish color cast, low red color, and low illumination in underwater images through contrast correction in both RGB and HSI color models\cite{iqbal2010enhancing}. 
	Hitam et al. combined contrast limited adaptive histogram equalization (CLAHE) significantly enhances underwater image quality by addressing visibility challenges through CLAHE applied on RGB and HSV color models\cite{hitam2013mixture}. 
	Li et al proposed a systematic underwater image enhancement method, integrating a dehazing algorithm and contrast enhancement, effectively restores visibility, color, and contrast, offering dual output versions for display and information extraction\cite{li2016underwater}.
	These traditional methods, whether model-based or non-model-based, have contributed to the field of UIE by addressing specific challenges and improving the visual quality of underwater images. However, they may have limitations in handling complex underwater scenarios and achieving robust and consistent enhancement results.}

{\textbf{Model-based deep learning UIE methods} often involve estimating unknown parameters in the underwater imaging model using convolutional neural networks (CNNs)\cite{li2021underwater,wang2021joint,8932601}. 
	For example, Li et al. developed the Ucolor network which effectively addresses color casts and low contrast in underwater images by leveraging multi-color space embedding and medium transmission-guided decoding\cite{li2021underwater}. 
	Wang et al. proposed a joint iterative network effectively addresses color cast and haze effects in underwater images through a novel triplet-based color correction module, recurrent dehazing module, and iterative optimization mechanism\cite{wang2021joint}. 
	Ye et al. Introduced an unsupervised adaptation network for joint underwater depth estimation and color correction. Utilizes style and feature-level adaptation to mitigate data scarcity and improve generalization. demonstrating effectiveness in addressing underwater vision challenges\cite{8932601}.
	However, the imprecise estimation of these parameters can limit the performance of these methods.}

{\textbf{End-to-end deep learning UIE methods}, where the input damaged image is directly passed through a deep learning model that outputs the enhanced image. 
	For instance, Wang et al. proposed an end-to-end framework, UIE-Net, addresses color distortion and visibility degradation in underwater images through a CNN-based approach trained for color correction and haze removal tasks simultaneously\cite{wang2017deep}.
	Jiang et al. presented TOPAL, a target-oriented perceptual adversarial fusion network. It features multi-scale contrast enhancement, color correction, dual attention fusion for detail preservation, and global-local adversarial training,  effectively enhances underwater image quality by addressing turbidity, chromatism, and detail preservation.
	Y. Wei et al. introduced UHD-SFNet model, with the aid of U-RSGNet and UHD-CycleGAN for paired dataset construction, efficiently addresses low contrast and blur in UHD underwater imaging by recovering color and texture in both frequency and spatial domains\cite{wei2022uhd}.
	Zhou et al. presented UGIF-Net, which enhances color information using two-color spaces, GI attention, dense blocks, and an MCA module, aiming to address real-world reference limitations and improve color utilization for richer results\cite{zhou2023ugif}.
	Jiang et al proposed LCNet, a Lightweight Cascaded Network based on Laplacian image pyramids, which effectively addresses the challenge of underwater image enhancement by progressively predicting high-quality residuals with reduced complexity\cite{10196309}.}

{These deep learning-based methods have shown promising results and the potential for improving underwater image quality. To our knowledge, existing UIE methods, even those considering enhancement in the frequency domain\cite{wei2022uhd,Huo_2021_ICCV}, have not deeply explored the relationship between frequency domain information and underwater image degradation factors. By decomposing images in the frequency domain, we decouple the factors contributing to image degradation, enabling separate enhancement of different aspects of the image. This is our major advantage compared to previous methods.}	
\begin{figure*}[b]
	\centering
	\includegraphics[width=0.9\textwidth]{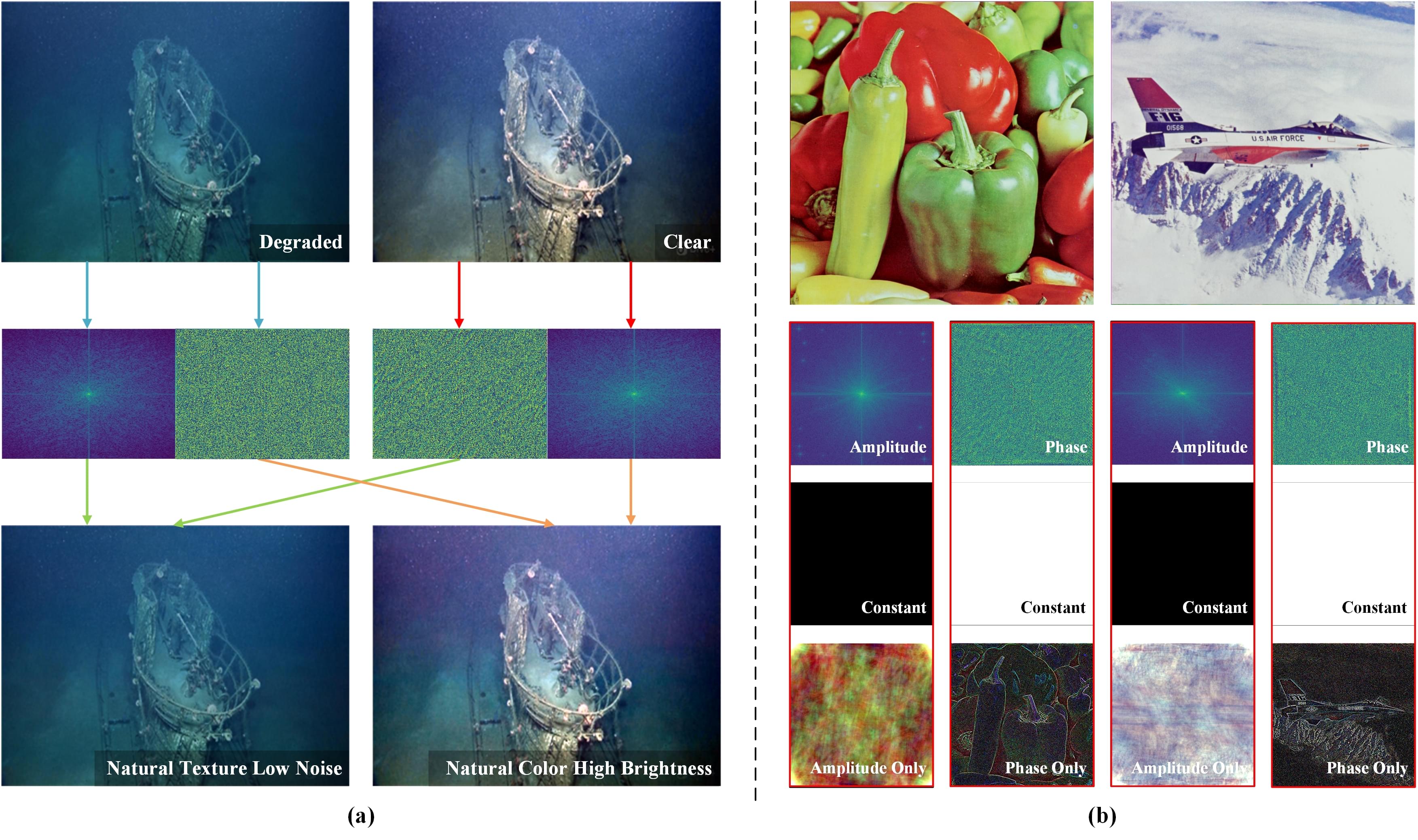}
	\caption{(a) By interchanging the amplitude and phase of the damaged and reference images in the frequency domain and reconstructing the images we find that the key factors causing underwater image degradation can be decoupled in the frequency domain where brightness and color information are expressed in amplitude while texture details and noise are reflected in phase. (b) Decomposing an image using FFT into phase and amplitude, then reconstructing images with either component reveals that phase mainly encodes texture and structure, while amplitude captures color and illumination details.}
	\label{fig:Observation}
\end{figure*}
\subsection{Transformer}
The Transformer architecture, introduced by Vaswani et al.\cite{vaswani2017attention}, has gained widespread popularity in natural language processing due to its ability to efficiently capture long-range dependencies and its large model capacity. Transformer-based models have achieved remarkable performance, even surpassing human-level performance in specific language tasks\cite{kenton2019bert}. The self-attention mechanism in Transformers is particularly effective in modeling long-range dependencies and capturing global features. Taking advantage of these strengths, Transformer models have also demonstrated excellent performance in various visual tasks\cite{dosovitskiy2020image,liu2021swin,zamir2022restormer,saleh_adaptive_2022,fan2024kmt}.
Query-based transformers are a variant of Transformer models specifically designed for processing and generating queries. These models utilize the Transformer architecture because of its ability to leverage attention mechanisms for capturing global correlations. Carion et al. introduced DETR (Detection Transformer)\cite{carion2020end} as an approach to redefine object detection by directly predicting sets based on queries. Instead of relying on a large number of region proposals, DETR employs only a small number of learned object queries (e.g., 100) as input.
Subsequent studies in object detection, such as PERFORMER\cite{cheng2021per} and Instances as Queries\cite{fang2021instances}, have further enhanced the query-based approach and achieved comparable performance to SOTA detectors like Cascade RCNN\cite{cai2018cascade}. These results indicate that the use of query-based transformers is a promising research direction in the field of object detection. As a result, there is a strong interest in extending the query-based framework to other scenarios and tasks.
In this paper, we propose the utilization of a color encoder to obtain a color embedding that is semantically aware. This is accomplished by employing an adaptive color query operation on image features at multiple scales. By employing this strategy, we aim to improve the performance of the model in handling semantic-dependent color recovery.

\section{Proposed Method}
In this section, we present the framework and details of FDCE-Net, which is designed to restore degraded underwater images and preserve texture details and color consistency. The goal is to map an input degraded underwater image $x \in \mathbb{R}^{H \times W \times C}$ to its corresponding normal-clear version $y \in \mathbb{R}^{H \times W \times C}$, where $H$, $W$, and $C$ represent the height, width, and number of channels, respectively. FDCE-Net has two major components (i.e., FS-Net, DCE) and one fusion net.

\subsection{Motivation}
In the initial analysis, we examine the underwater degraded image and the reference image. By decomposing both the degraded image and the enhanced image into their amplitude and phase components using the Fast Fourier Transform (FFT), and then performing the Inverse Fast Fourier Transform (IFFT) with the exchange of amplitude and phase, we observe the reconstructed image shown in Fig. \ref{fig:Observation}(a). From the observation, we notice that combining the amplitude of the reference image with the phase of the degraded image yields an image with normal brightness and color. However, this combination introduces additional artifacts and noise. On the other hand, combining the phase of the reference image with the amplitude of the degraded image produces an image with more natural texture details and less noise.
{Further investigation validated our hypothesis, as illustrated in Fig. \ref{fig:Observation}(b), where we decompose a random image into its phase and amplitude components using FFT. Subsequently, we combined its phase with a constant amplitude that lacks meaningful content to form a new image through IFFT. It is observed that the image reconstructed solely from the phase predominantly contains texture and structural information, underscoring the phase's role in encoding the image's spatial information. Similarly, merging its amplitude with a constant phase, which also carries no significant information, into a new image by IFFT, reveals that an image rebuilt only from amplitude mainly contains color and illumination details, with the texture and structure being entirely absent.}

On the basis this observation, we conclude that we can to some extent separate the key factors affecting the degradation of underwater images in the frequency domain. The brightness and color information are predominantly expressed in the amplitude, while the texture and noise characteristics are reflected in the phase. Leveraging this prior knowledge, we design a FS-Net that incorporates FSRB for the initial restoration of the image. The specific details of this network will be further explained in subsection C.

\begin{figure*}[t]
	\centering
	\includegraphics[width=0.95\textwidth]{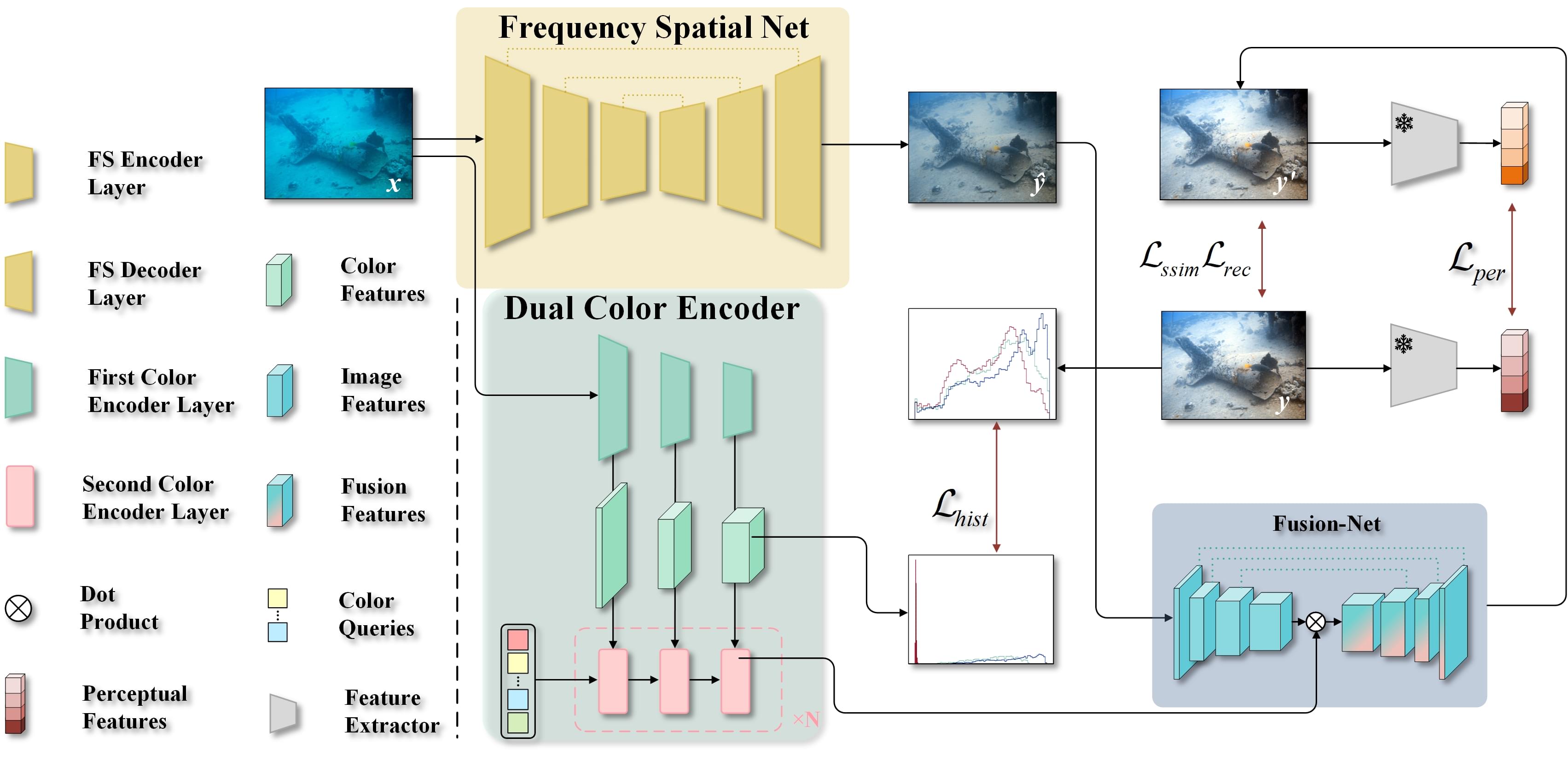}
	\caption{Overview of the proposed FDCE-Net architecture, which comprises two main components: FS-Net and DCE, enhances an underwater image $x$ in an end-to-end fashion. Initially, $x$ is simultaneously inputted into FS-Net and DCE, where FS-Net provides preliminary enhancement to produce $\hat{y}$. Concurrently, the first color encoder extracts its multi-scale features, followed by the second color encoder performing color queries on visual features at different scales to learn semantic-aware color representations. The Fusion-Net combines the outputs of FS-Net and DCE to generate a well-enhanced image $y'$.}
	\label{fig:Overview}
\end{figure*}

\subsection{Overview}
Based on the aforementioned observations, we propose the use of FS-Net to initially enhance the input image by incorporating color and texture details. The FS-Net takes the input image $x$ and first embeds it into the feature domain through a convolutional layer. These features are then processed by an encoder-decoder network, where the main computation is performed by Frequency Spatial Residual Blocks (FSRB). The FSRB includes three 2× downsampling operations and three 2× upsampling operations. Each upsampling layer in the FSRB has a shortcut connection to the corresponding stage of the encoder. The output of FS-Net is the coarse result denoted as $\hat{y}$.

Simultaneously, we also feed the original image $x$ into the DCE. DCE starts with a color encoder network, which also consists of FSRB as its computational blocks. The first color encoder network extracts visual features from the image. These features are then passed to the second image encoder, which generates semantically aware color embeddings by performing adaptive color query operations on image features at different scales.

Finally, the Fusion-Net combines the initial enhanced image output from FS-Net with the color features generated by DCE. This fusion process produces a vivid and semantically aware output image $y'$.

For the pseudo-code representation of our FDCE-Net, please refer to Algorithm \ref{algo1}.

\begin{algorithm}[h]
	\SetAlgoLined
	\KwIn{Degraded Underwater Image $x$}
	\KwOut{Clear Enhanced Image $y$}
	$\hat{y} \leftarrow \text{FS-Net}(x')$\;
	$CF_1, CF_2, CF_3 \leftarrow FCE(x)$\;
	Zero initialize $E$\;
	
	\For{$i\leftarrow 1$ \KwTo $N$}{
		\For{$j\leftarrow 1$ \KwTo $3$}{
			$E \leftarrow CEB_j(E, CF_j)$\;
		}
	}
	$IF \leftarrow FusionNetEncoder(\hat{y})$\;
	$FF \leftarrow IF \cdot E$\;
	$y' \leftarrow FusionNetDecoder(F)$\;
	
	\caption{Workflow of the proposed FDCE-Net}
	\label{algo1}
\end{algorithm}

\subsection{Frequency Spatial Network}
{Most existing UIE methods predominantly rely on spatial information, overlooking the significance of frequency information. Although some methods \cite{wei2022uhd, 9607639} have incorporated frequency information for underwater image enhancement, they typically treat it merely as an additional feature, thus undervaluing its potential in deciphering degradation factors within underwater images.}
{In our motivation, we noted that attributes such as color, luminance, texture, and noise characteristics can be effectively decomposed in the frequency domain. Consequently, the primary objective of FS-Net is to leverage this observation to predict an initially enhanced image. This enhanced image prominently features improved color fidelity, increased luminance, minimal noise, and enriched texture details. The design of our basic processing block in FS-Net adopted a dual paths structure, it has a frequency path and a spatial path, Although our observation is found in the frequency domain, the use of spatial domain information is necessary. Information from the frequency domain and the spatial domain represent different forms of feature expression. Frequency domain information is commonly used to capture the global structure and texture information of images, while spatial domain information focuses more on local details and features such as edges. In FS-Net, we employ an encoder-decoder pipeline that resembles the traditional U-Net\cite{Ronneberger2015unet} structure. However, instead of using basic convolution or Residual Blocks as the main computational unit, we introduce the Frequency Spatial Residual Block (FSRB) to achieve amplitude and phase enhancement in the frequency domain and feature enhancement in the spatial domain simultaneously.}

\begin{figure*}[t]
	\centering
	\includegraphics[width=1\textwidth]{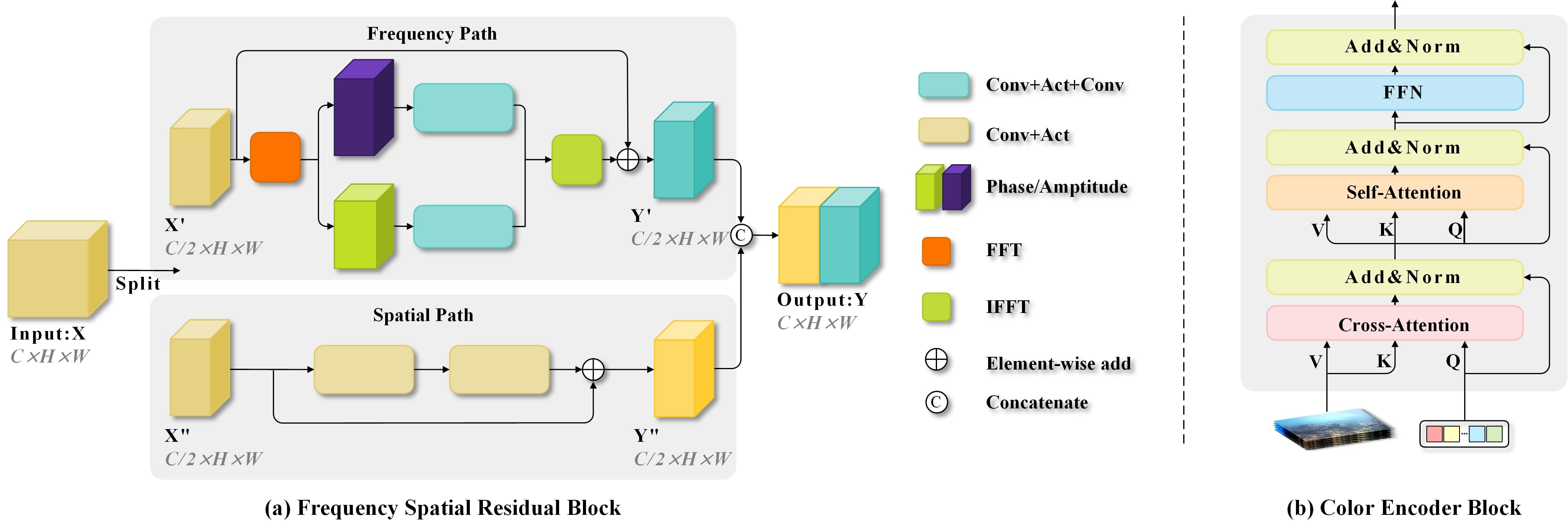}
	\caption{(a) Schematic illustration of the proposed Frequency Spatial Residual Block
		(b) The design of the transformer-based color encoder block involves taking image features and trainable color queries as input and establishing a connection between semantic and color representation through cross-attention, self-attention, and feed-forward operations.}
	\label{fig:FSRB_CEB}
\end{figure*}

{As illustrated in \ref{fig:FSRB_CEB}, the input features X are split into two parts, X' and X''. X' is fed into the frequency path. In this path, we first apply the Fast Fourier Transform (FFT) to obtain the amplitude component and the phase component. These components are then separately passed through two convolutional layers with a 1×1 kernel size. We use 1x1 convolution kernels to prevent large kernels from disrupting the structural information of phase and amplitude. We use the LeakyReLU activation function in between. The separate processing of amplitude and phase achieves enhancement of different aspects of the image. Afterward, we transform them back to the spatial domain by applying the Inverse Fast Fourier Transform (IFFT) and combine them with X' in a residual manner to obtain Y'.}

{In parallel, we feed X'' into the spatial path. Two 3×3 convolutional layers with the GELU activation function are applied, and the result is combined with X'' in a residual manner to obtain Y''. 
	Finally, we concatenate Y' and Y'' together to obtain the final output feature Y. By integrating these two types of information, the network can obtain rich feature representations from different perspectives, thereby enhancing the model's expressive capabilities and facilitating the decoupling of degradation factors\cite{10411925}.}

\subsection{Dual Color Encoder}
{Color attenuation is another significant challenge in underwater imagery. Previous studies\cite{yang2011low, peng2015single, song2018rapid, li2020underwater} have introduced various priors to address this issue. For example, some methods utilize the Dark Channel Prior (DCP)\cite{yang2011low, peng2015single, liang2021gudcp}, while others are based on underwater scene priors\cite{song2018rapid, li2020underwater} or histogram distribution priors\cite{li2016underwater}; however, due to the complex underwater environment and variations in depth at which images are captured, the extent and pattern of color attenuation can vary. Consequently, a single prior may not be suitable for all scenarios. To overcome the problem of color shift/attenuation. To reduce the dependence on manually designed priors, we propose the use of two color encoders, known as Dual Color Encoders. By leveraging the correlation between color and semantic representations
	via cross-attention, it learns a set of adaptive color embeddings based on visual semantic information,
	eliminating reliance on manually crafted priors and enabling generalization across diverse conditions.}

{\textbf{The first color encoder} is responsible for extracting color information and semantic embeddings of the image at different scales. It consists of multiple sets of convolutional and downsampling operations that gradually reduce the image resolution, aiming to obtain multi-scale semantic embeddings and color features from the image. The first color encoder captures a complete image feature pyramid through a step-by-step downsampling process. These multi-scale features are further utilized as input to the second color encoder to guide the optimization of color queries. Additionally, we utilize the first color encoder to compute the color histogram of the image in the RGB color space. A color histogram represents the distribution of colors within an image by tallying the occurrences of specific pixel intensities or ranges of pixel intensities. During the training process, we enforce the first color encoder to acquire consistent color features with the reference image. However, as the color histogram describes color information globally, it lacks spatial information and cannot provide suitable color information for specific image contents. Consequently, the direct connection between color and content features cannot be established. This is why we introduce the second color encoder, which bridges the gap between color queries and semantic information.}

\textbf{The second color encoder} consists of a sequence of blocks, each of which receives a visual feature and a color query as input. These color encoder blocks are based on a modified Transformer module, as depicted in Fig. \ref{fig:FSRB_CEB}(b). The objective of the color encoder is to acquire a collection of adaptive color embeddings derived from visual semantic information. Consequently, we establish trainable color embedding repositories for storing sequences of color representations:
${E_0}=[E_0^1,E_0^1,...,E_0^M]\in{{\mathbb{R}}^{M \times C}}$
where M represents sequence length, C refers to color channels. In the training stage, these color embeddings are initially set to zero. Subsequently, they serve as color queries in the first encoder block. 

To begin, we build the correlation between semantic representations and color embeddings using a cross-attention layer: 
\begin{equation}
	{}^1E_l = softmax(Q_lK_l^T)V_l+E_{l-1},
\end{equation}
the layer index is denoted by $l$, and the M C-dimensional color embeddings of layer $l$ is represented by $E_l\in {{\mathbb{R}}^{M \times C}}$. The color query $Q_l\in{{\mathbb{R}}^{M \times C}}$ of layer $l$ is obtained by linearly transforming the color embedding $E_{l-1}$ of layer $l-1$. The image features are map by two separate linear transformation to $K_l,V_l\in{{\mathbb{R}}^{H_l \times W_l \times C}}$, respectively. $H_l$ and $W_l$ represent the height and width of image features. Image features are used to enhance the color embedding by cross-attention operation. The color embedding is then transformed using the standard transformer layer as follows: 
\begin{equation}
	{}^2E_l = MHSA(LN({}^1{E_l})) + {}^1{E_l},
\end{equation}
\begin{equation}
	E_l = LN(FFN(LN({}^2{E_l})) + {}^2{E_l}),
\end{equation}
where $MHSA( \cdot )$ refers to multi-head self-attention, $FFN( \cdot )$ denotes feed-forward network, and $LN( \cdot )$ stands for layer normalization. In our color encoder module, cross-attention is performed before self-attention, this is because color queries are zero-initialized and semantically independent before entering the first self-attention layer. It is the main difference between our color encoder block and the traditional transformer module. 

We grouped the blocks into 3 encoder blocks each group. The intermediate visual features generated by the first color encoder using downsampling rates of 1/2, 1/4, and 1/8 are sequentially fed to these encoder blocks. We repeat the group N times in a cyclic manner. The color encoder consists of a total of 3N encoder blocks. We formulate the color encoder as follows: 
\begin{equation}
	E = ColorEncoder(E_0, F_1, F_2, F_3),
\end{equation}
where $F1,F2,F3$ are three different scale of visual features. In the second color encoder, a combination of multi-scale features is employed to capture the connection between the color embedding and the visual embedding, thereby enhancing the color embeddings' sensitivity to semantic content.

\subsection{Fusion Network}
Fusion-Net and FS-Net indeed have a shared structure. In Fusion-Net, the preliminary enhanced images from the output of FS-Net and the color embeddings from the DCE are aggregated. The preliminary enhanced images are directly fed into Fusion-Net for further processing.
The fusion module in Fusion-Net combines the features of the initially enhanced images and the color embeddings using a simple dot product operation
\begin{equation}
	FF = IF \cdot E.
\end{equation}
This fusion process generates aggregated features that capture both the enhanced image information and the semantic color information. The aggregated features are then passed through the decoder of the fusion network, which generates the final fully enhanced images.

By integrating the preliminary enhanced images and color embeddings, Fusion-Net benefits from both the initial enhancement information and the semantic color information. This fusion mechanism enhances the overall image quality and preserves the desired color characteristics in the final enhanced images.

\begin{equation}
	y' = FusionNetDecoder(FF).
\end{equation}

\subsection{Loss Function}
To ensure the robustness of the model, it is crucial to define a suitable loss function that quantifies the consistency between the model's predictions and the reference values in the training dataset. Let's consider the reference image as $y$, and denote the degraded image and predicted image as $x$ and $y'$, respectively.

For assessing the effectiveness of the predicted images in the spatial domain, we incorporate a loss function based on structural similarity to ensure a favorable visual perception outcome during the network learning process.

The structural similarity index (SSIM) is widely used to measure the perceptual similarity between two images. It takes into account not only pixel-wise differences but also considers the structural information and the correlation between local image patches. The mathematical expression of the SSIM loss function is as follows:
\begin{equation}
	\mathcal{L}_{ssim}=1 - {\left( {\frac{{2{\mu _e}{\mu _r} + {c_1}}}{{\mu _e^2 + \mu _r^2 + {c_1}}}} \right)^\alpha }{\left( {\frac{{2{\sigma _{er}} + {c_2}}}{{\sigma _e^2 + \sigma _r^2 + {c_2}}}} \right)^\beta },
\end{equation}
where $\mu _e$ and $\mu _r$ denote the respective mean value of predicted image and reference image. $\sigma _e$ and $\sigma _r$ correspond to the standard deviation of the predicted image and the clear image, while $\sigma _{er}$ represents the covariance between the predicted and clear values. Additionally, $\alpha$ and $\beta$ signify the relative significance of the two components, and c1 and c2 are constants.

In addition to the SSIM loss for supervision, we also employ the L1 loss as a reconstruction loss. The L1 loss has been widely used in UIE tasks and is mathematically represented as follows:
\begin{equation}
	\mathcal{L}_{rec}=\frac{1}{N}\sum\limits_{i = 1}^N {{{\left\| {{y_i}' - y_i} \right\|}_1}},
\end{equation}
where $N$ is the number of image's pixel, and $i$ is the index of the pixel.

The color histogram is a widely used color feature in image retrieval systems\cite{pass1996histogram,deng2001efficient}. To enhance the fidelity of color histogram reconstruction, we introduce a constraint based on the L1 distance between the histogram generated by the first color encoder and the histogram of the reference image. The constraint can be defined as follows:
\begin{equation}
	\mathcal{L}_{hist}=\frac{1}{{C \times bin}}{\left\| {{H_{fce}} - {H_y}} \right\|_1},
\end{equation}
where  $H_{fce}$ represents color histogram calculated by the first color encoder, $H_y$ denotes the actual color histogram of the reference image, $C=3$ corresponds to the three color channels, $bin$ refers to the number of bins we use when calculating color histogram.

In addition to the previously mentioned loss functions, we incorporate a perceptual loss function by utilizing a pre-trained VGG16 model, which was trained on the ImageNet dataset, to quantify perceptual similarity. The perceptual loss function can be formulated as follows:
\begin{equation}
	\mathcal{L}_{per}=\mathop \sum \limits_{j = 1}^3 \frac{1}{{{H_j}\times{W_j}\times{C_j}}}\parallel {\Phi _j}\left( {y'} \right) - {\Phi _j}\left( y \right)\parallel _2^2,
\end{equation}
where $H_j$, $W_j$, and $C_j$ denote the dimensions of the feature map at layer $j$ of the network, signifying its height, width, and number of channels, respectively. The function ${\Phi _j}( \cdot )$ denotes the activation function applied at layer $j$.

The total loss function used to guide the training of our proposed approach is a combination of the previously mentioned loss functions. It can be depicted as follows:
\begin{equation}
	\mathcal{L}_{total}=\mathcal{L}_{ssim}+\mathcal{L}_{rec}+\alpha\mathcal{L}_{hist}+\beta\mathcal{L}_{per},
\end{equation}
where $\alpha$ and $\beta$ representing the hyperparameters associated with each individual loss component.

\begin{table*}[htbp]
	\renewcommand{\arraystretch}{1.2}
	\centering
	\caption{{Quantitative results of UIE on the UIEB, EUVP and UIQS datasets as well as computational complexity, we use SSIM↑, PSNR↑, LPIPS↓ and MSE ($\times10^3$)↓ for paired datasets and UIQM↑, UCIQE↑ and UIF↑ for unpaired datasets, where ↑ larger values mean better results and ↓ denotes that smaller values mean better results. The \textbf{best} and \underline{second} results are shown in boldface and underscore.}}
	\resizebox{\linewidth}{!}{
		\begin{tabular}{cccccccccccc}
			\toprule[1pt]
			\multicolumn{1}{c|}{\multirow{2}[2]{*}{Dataset}} & \multicolumn{1}{c|}{\multirow{2}[2]{*}{Metric}} & \multicolumn{1}{c|}{UWCNN} & \multicolumn{1}{c|}{FUnIE-GAN} & \multicolumn{1}{c|}{Ucolor} & \multicolumn{1}{c|}{UHDSF-Net} & \multicolumn{1}{c|}{PUIE} & \multicolumn{1}{c|}{MMLE} & \multicolumn{1}{c|}{TACL} & \multicolumn{1}{c|}{Semi-UIR} & \multicolumn{1}{c|}{WWPE} & \multirow{2}[2]{*}{Ours} \\
			\multicolumn{1}{c|}{} & \multicolumn{1}{c|}{} & \multicolumn{1}{c|}{20'PR} & \multicolumn{1}{c|}{20'RA-L} & \multicolumn{1}{c|}{21'TIP} & \multicolumn{1}{c|}{22'ACCV} & \multicolumn{1}{c|}{22'ECCV} & \multicolumn{1}{c|}{22'TIP} & \multicolumn{1}{c|}{22'TIP} & \multicolumn{1}{c|}{23'CVPR} & \multicolumn{1}{c|}{23'TCSVT} &  \\
			\addlinespace[2pt]
			\hline
			\hline
			\multirow{4}[2]{*}{UIEB} & SSIM  & 0.652 & 0.738 & 0.888 & \underline{0.897} & 0.891 & 0.755 & 0.852 & 0.884 & 0.767 & \textbf{0.917} \\
			& PSNR  & 14.08 & 17.84 & 22.56 & 22.56 & 21.85 & 17.02 & \underline{22.76} & 22.48 & 17.97 & \textbf{23.87} \\
			& LPIPS & 0.262 & 0.241 & \underline{0.077} & 0.107 & 0.088 & 0.240 & 0.107 & 0.090 & 0.186 & \textbf{0.074} \\
			& MSE   & 3.234 & 1.457 & \underline{0.466} & 0.574 & 0.594 & 1.650 & 0.497 & 0.548 & 1.289 & \textbf{0.419} \\
			\hline
			\addlinespace[1pt]
			\multirow{4}[2]{*}{EUVP} & SSIM  & 0.719 & 0.831 & 0.828 & \underline{0.887} & 0.816 & 0.643 & 0.779 & 0.797 & 0.656 & \textbf{0.893} \\
			& PSNR  & 17.22 & 23.04 & 20.58 & \underline{26.11} & 19.35 & 15.01 & 19.05 & 19.13 & 15.89 & \textbf{26.17} \\
			& LPIPS & 0.277 & 0.216 & 0.244 & 0.221 & 0.249 & 0.253 & \underline{0.204} & 0.238 & 0.287 & \textbf{0.110} \\
			& MSE   & 1.578 & 0.378 & 0.672 & \textbf{0.146} & 0.884 & 2.507 & 0.992 & 0.934 & 2.011 & \underline{0.185} \\
			\hline
			\addlinespace[1pt]
			\multirow{4}[1]{*}{LSUI} & SSIM  & 0.729 & 0.837 & 0.848 & \underline{0.856} & 0.848 & 0.696 & 0.813 & 0.827 & 0.700 & \textbf{0.881} \\
			& PSNR  & 17.41 & \underline{22.63} & 21.39 & 22.41 & 20.89 & 16.86 & 19.92 & 20.47 & 17.12 & \textbf{24.47} \\
			& LPIPS & 0.263 & 0.218 & 0.211 & 0.217 & 0.199 & 0.234 & \underline{0.183} & 0.205 & 0.254 & \textbf{0.154} \\
			& MSE   & 1.556 & 0.458 & 0.576 & \underline{0.360} & 0.736 & 1.627 & 0.841 & 0.806 & 1.502 & \textbf{0.288} \\
			\hline
			\addlinespace[1pt]
			\multirow{3}[1]{*}{UIQS} & UIQM  & 3.155 & 3.954 & 3.603 & 3.640 & 3.746 & 4.140 & \textbf{4.339} & 3.846 & 4.232 & \underline{4.253} \\
			& UCIQE & 0.467 & 0.522 & 0.530 & 0.531 & 0.541 & 0.579 & 0.592 & 0.567 & \underline{0.593} & \textbf{0.598} \\
			& UIF   & 0.603 & 0.361 & 0.520 & \underline{0.679} & 0.605 & 0.311 & 0.392 & 0.602 & 0.435 & \textbf{0.705} \\
			\hline
			\addlinespace[1pt]
			\multicolumn{2}{c}{GFLOPs} & 3.08 & 10.23 & 443.85 & 14.58 & 423.05 & \textbackslash{} & \textbackslash{} & 66.2 & \textbackslash{} & 12.78 \\
			\multicolumn{2}{c}{Params (M)} & 0.04 & 7.019 & 157.4 & 29.76 & 1.40 & \textbackslash{} & 22.76 & 65.6 & \textbackslash{} & 5.53 \\
			\multicolumn{2}{c}{Inference Time (s)} & 1.43 & 0.14 & 2.75 & 0.11 & 0.02 & 0.08 & 0.27 & 0.08 & 0.30 & 0.09 \\
			\bottomrule[1pt]
			
		\end{tabular}%
	}
	\label{tab:SOTA_Comparison}%
\end{table*}%

\section{Experiments}
In this section, we commence by presenting a comprehensive overview of the datasets utilized in our study. Subsequently, we delve into the details of our experimental configurations and the criteria we employ for evaluation. Additionally, we conduct a thorough quantitative and qualitative comparison of our proposed method with SOTA UIE methods. Moreover, we perform an ablation study on our network architecture to assess the impact of various components and modules.
\subsection{UIE Datasets}
\textbf{UIEB (Underwater Image Enhancement Benchmark)}\cite{li2019underwater} is a dataset widely used for the UIE task, which contains 890 raw underwater images with corresponding high-quality reference images, we randomly select a subset of 790 images as the training set and the remaining 100 as the test set.

\textbf{EUVP (Enhancing Underwater Visual Perception)}\cite{islam2020fast} dataset includes distinct collections of both paired and unpaired image samples, representing scenarios with varying degrees of perceptual quality. This dataset has been curated to support the supervised training of models designed for enhancing underwater image quality. We use the 2185 images in the Underwater Scenes subset, which are randomly split into a training set containing 1985 images and a test set of 200 images.

{\textbf{LSUI (large-scale underwater image)}\cite{10129222} dataset, which covers abundant underwater scenes and visual quality reference images. The dataset contains 4279 real-world underwater image groups, in which each raw image’s clear reference images are paired correspondingly. We randomly partitioned 3779 pairs for the training set and 500 pairs for the test set.}

\textbf{UIQS (Underwater Image Quality Set)}\cite{liu2020real} unpaired dataset is a subset of RUIE (Real-world Underwater Image Enhancement). RUIE has three subdivisions: UIQS, UCCS (Underwater Color Cast Set), and UHTS (Underwater Higher-level Task-driven Set), each addressing one of the three demanding aspects or objectives related to enhancement, namely, degradation in visibility, color distortion, and advanced-level detection/classification, respectively. We randomly selected 365 images in the UIQS subset to test the effectiveness of the pre-trained model on UIEB datasets.

\begin{figure*}[h]
	\centering
	\includegraphics[width=0.98\textwidth]{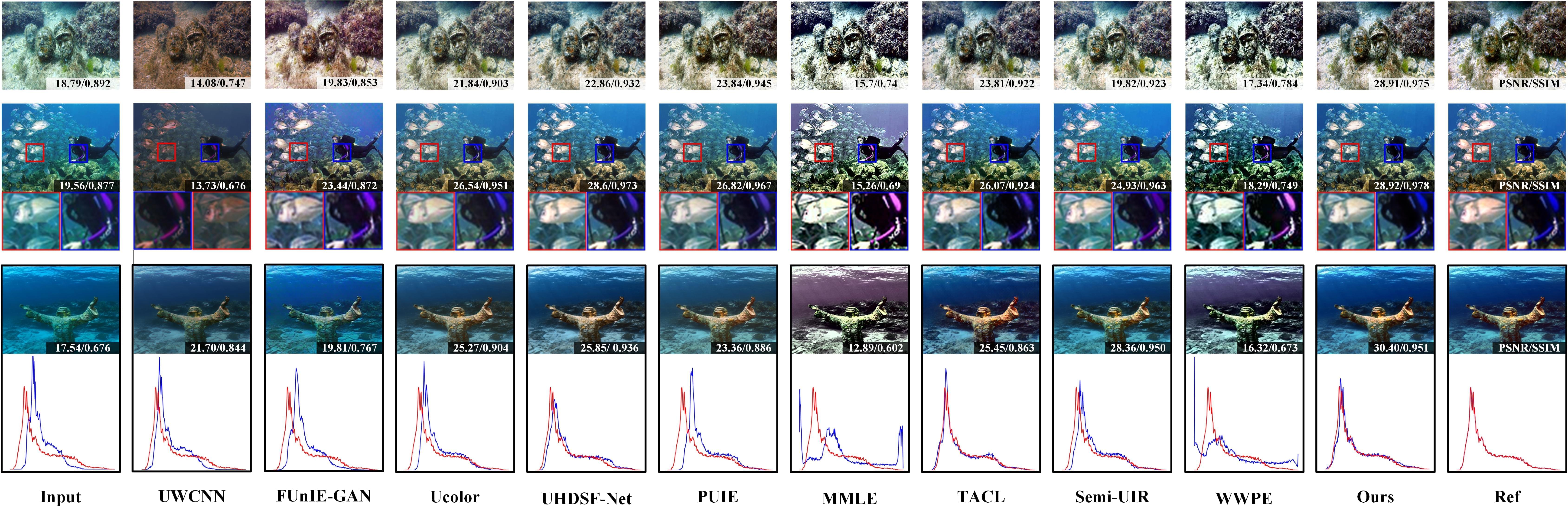}
	\caption{Qualitative comparison of UIEB, the first row is used to compare the overall enhancement effect, the second row of images zooms in to compare some of the details in the images, and the last row of images compares the pixel distribution graph of the image, the red line represents the pixel distribution of the reference image and the blue line represents the pixel distribution of the enhanced image.}
	\label{fig:UIEB}
\end{figure*}

\begin{figure*}[h]
	\centering
	\includegraphics[width=1\textwidth]{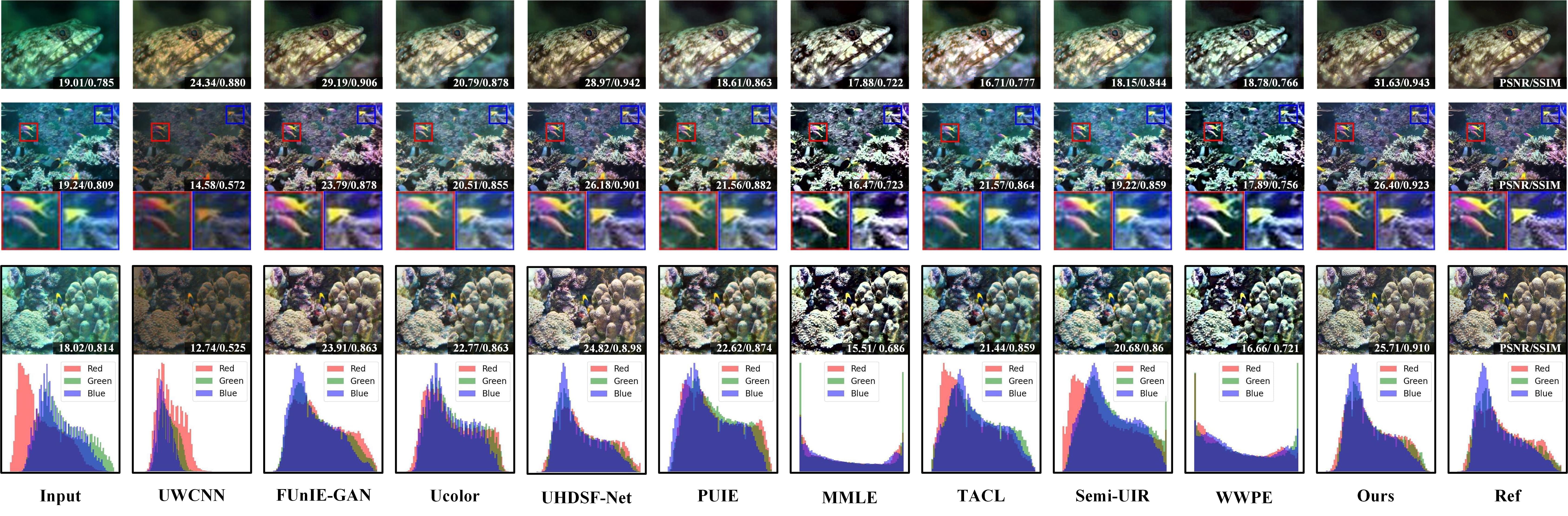}
	\caption{Qualitative comparison of EUVP, the first two rows of images are used to compare the overall enhancement effect, the second row of images zooms in to compare some of the details in the graph, and the last row of images compares the three-channel color histogram of the image.}
	\label{fig:EUVP}
\end{figure*}

\begin{figure*}[h]
	\centering
	\includegraphics[width=1\textwidth]{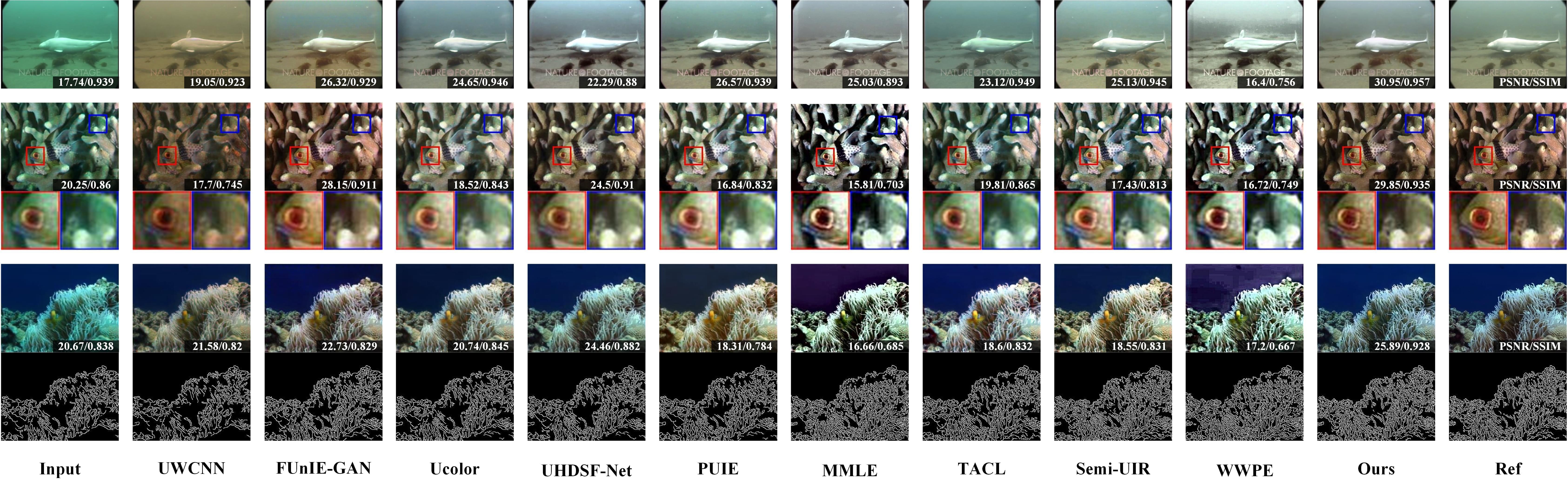}
	\caption{Qualitative comparison of LSUI, the first row is used to compare the overall enhancement effect, the second row of images zooms in to compare some of the details in the graph. The edge detection algorithm was applied to the third-row images to validate the enhancement of texture details.}
	\label{fig:LSUI}
\end{figure*}

\begin{figure*}[h]
	\centering
	\includegraphics[width=1\textwidth]{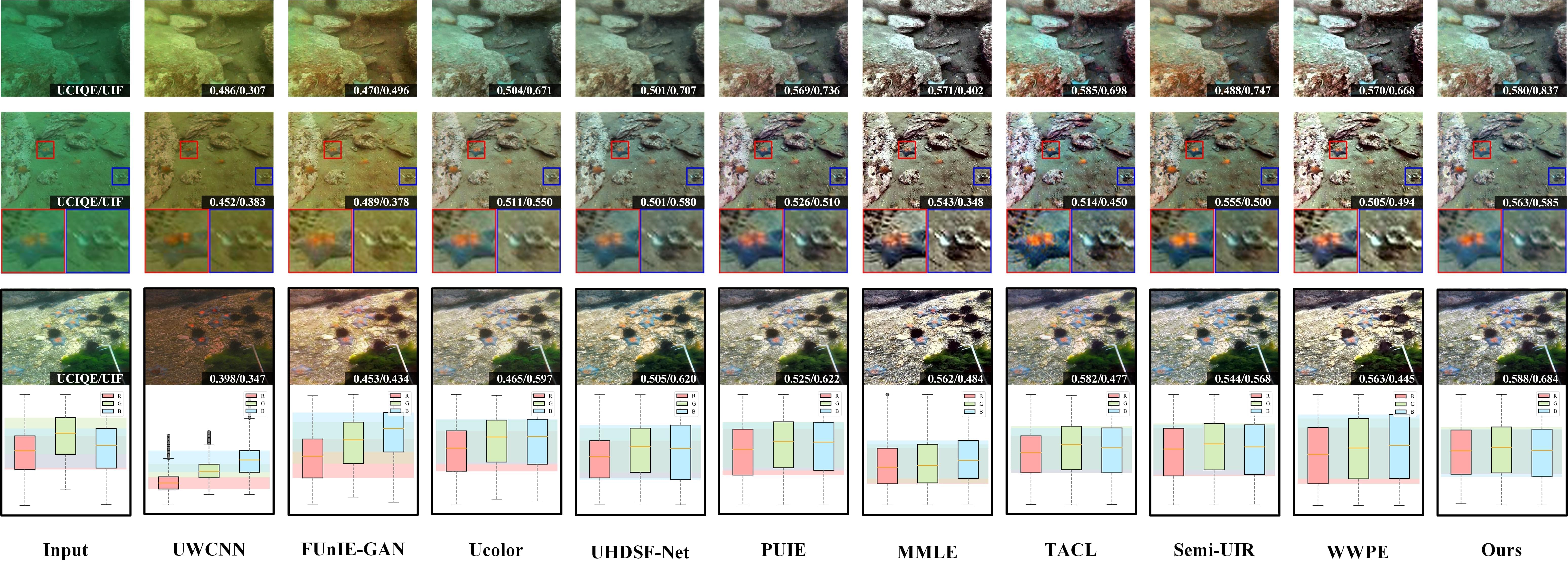}
	\caption{Qualitative comparison of UIQS, the first two rows of images are used to compare the overall enhancement effect, the second row of images zooms in to compare some of the details in the graph, and the last row of images compares the dispersion of RGB color space of the image.}
	\label{fig:UIQS}
\end{figure*}

\subsection{Experiment Settings}
To introduce diversity into our training dataset and enhance the robustness of our model, we incorporate random rotations of 90, 180, or 270 degrees, as well as vertical and horizontal flips. The optimization process is conducted using the AdamW optimizer, with default settings for $\beta_1$ and $\beta_2$, which are assigned values of 0.9 and 0.999, respectively. Other parameters adhere to the default settings in PyTorch. Our experiments are executed on a single RTX 3090 GPU. The model is trained for 40 epochs, with the learning rate gradually decreasing from $10^{-4}$ to $10^{-6}$. For the hyperparameters of our FDCE-Net, we empirically set $\alpha$ to 0.5 and $\beta$ to 0.05.

In order to validate the effectiveness of our proposed method, we compare the quantity and quality of our experimental results with SOTA methods, including UWCNN\cite{li2020underwater}, FUnIE-GAN\cite{islam2020fast}, Ucolor\cite{li2021underwater}, UHDSF-Net\cite{wei2022uhd}, PUIE\cite{Fu_2022}, MMLE\cite{9788535}, TACL\cite{liu2022twin}, Semi\-UIR\cite{huang2023contrastive}, WWPE\cite{10196309}. All learning-based methods are trained using the same training datasets, and their training methodologies align with the descriptions provided in their respective papers.

\subsection{Quantitative Evaluation}
\textbf{UIE dataset with paired data.} We conducted evaluations of paired datasets UIEB, EUVP and LSUI. First three rows of Tab. \ref{tab:SOTA_Comparison} presents the average scores for SSIM, PSNR, LPIPS and MSE ($\times10^3$). Firstly, it is evident that our method achieves significantly higher average scores compared to other methods. We outperform existing SOTA methods in almost all of the reference metrics. On the metrics of SSIM and PSNR on the UIEB dataset, our method improves 2.23\% and 4.88\%, respectively, over the best method currently available. For these two metrics on the LSUI dataset, our method improves 2.92\% and 8.13\%. Moreover, our method exhibits strong generalization capabilities on paired datasets.

\textbf{UIE dataset without paired data.} Furthermore, we conducted experiments on un-paired datasets UIQS, with the results presented in the last row of Tab. \ref{tab:SOTA_Comparison}. Upon reviewing the result, our method secured first place on UICQE as well as UIF\cite{zheng2022uif} and second place on UIQM. Taken jointly, our method performs more balanced and overall outperforms the SOTA method.

\subsection{Qualitative Evaluation}
\textbf{UIE dataset with paired data.} 
From the first row of Fig. \ref{fig:UIEB}, Fig. \ref{fig:EUVP} and Fig. \ref{fig:LSUI}, it is evident that our proposed FDCE-Net produces an overall enhancement effect for underwater images that is closer to the reference image. In comparison, the UWCNN-enhanced image appears generally dark and reddish, the FUnIE-GAN-generated image sometimes exhibits regular artifacts, and MMLE and WWPE introduce unnatural color in the enhanced images. Moreover, in the first image of Fig. \ref{fig:EUVP}, PUIE and TACL over-brighten the image, resulting in an unnatural appearance. Upon examining the zoomed-in details of the third image in Fig. \ref{fig:UIEB} and Fig. \ref{fig:EUVP}, it is evident that the restoration of the color of the diver's arms is most faithful to the reference image in our approach, followed by Ucolor, while the other methods failed to reproduce the correct colors. From the zoomed-in details of the second row images in Fig. \ref{fig:EUVP} and Fig. \ref{fig:LSUI}, it can be observed that our method outperforms other methods in both detail and color restoration. By comparing the pixel distribution graph of the last image in Fig. \ref{fig:UIEB} with the reference image, our results exhibit the closest resemblance in terms of distribution. Similarly, by comparing the three-channel color histogram of the last image in Fig. \ref{fig:EUVP} with the reference image, it is apparent that our outcomes align most closely with the RGB color space distribution. 
{Finally, comparing the edge detection results of Fig. \ref{fig:LSUI}, we can see that the edge detection results of the image enhanced using our method are closer to the reference image. Regions A and B exhibit more texture than the reference image, indicating the introduction of unreasonable texture during the image enhancement process. The remaining images show insufficient recovery of texture details.}
Overall, our method provides superior enhancement in terms of color and details, while other methods often suffer from drawbacks such as incorrect color recovery and missing details.

\textbf{UIE dataset without paired data.} 
In Fig. \ref{fig:UIQS}, we present a comparison of results on the UIQS dataset. Both FUnIE-GAN and UWCNN demonstrate insufficient enhancement, as the images still exhibit a greenish appearance. Upon examining the enlarged patches in the third row, it is evident that Ucolor and SC-Net lack details in the enhanced images, resulting in a smeared appearance. Conversely, TACL's image enhancement is over-enhanced with excessive contrast. The remaining methods generally provide a satisfactory overall impression. However, when analyzing the box plots for comparison, it becomes apparent that our proposed approach better addresses the challenge of varying degrees of underwater light attenuation.

\begin{figure}[t]
	\centering
	\includegraphics[width=0.48\textwidth]{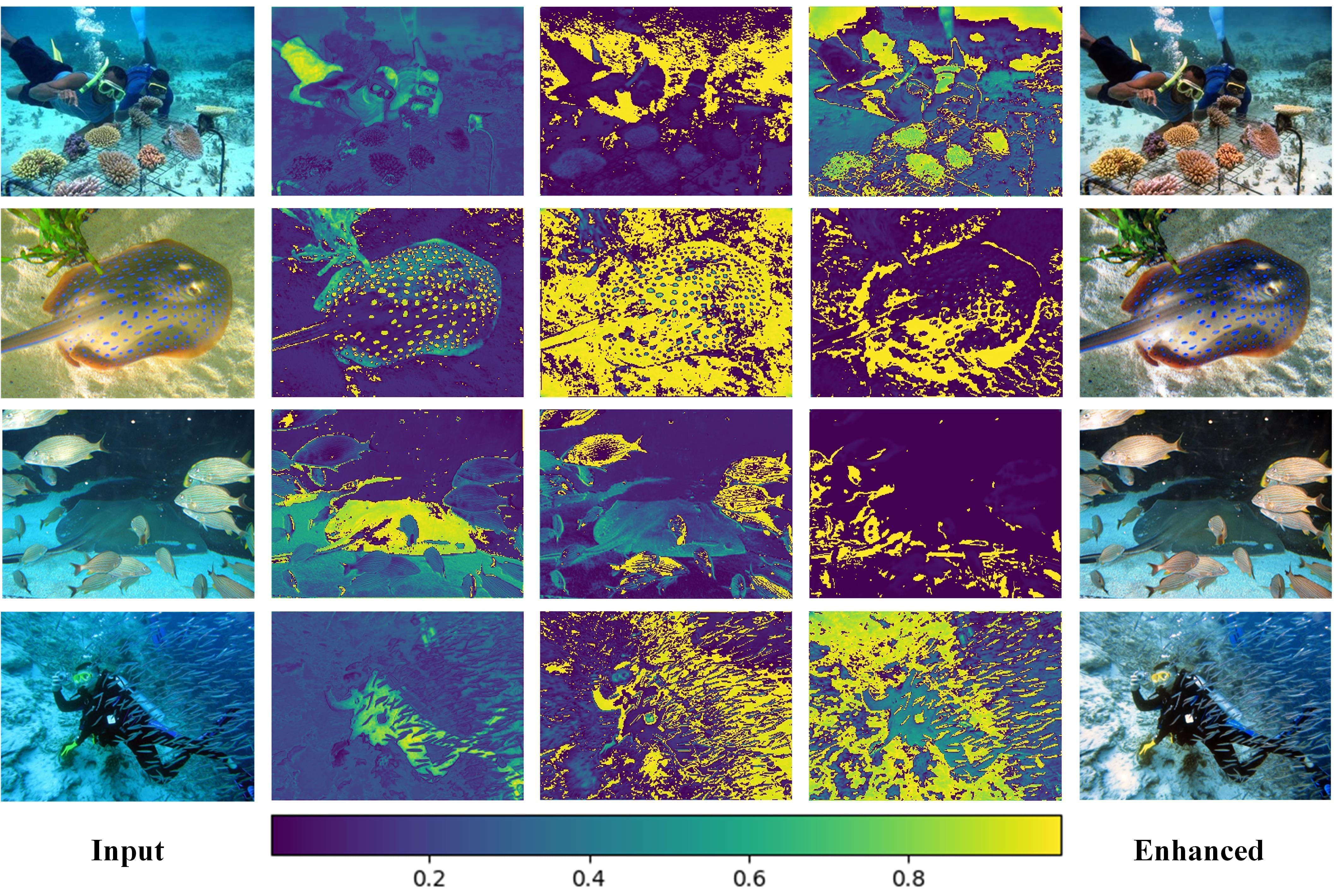}
	\caption{Visualization of the learned color query. The leftmost column is the input underwater damaged image, the rightmost is the image enhanced using our method, and the middle three columns visualize the color query. Yellow (purple) represents high (low) activation values}
	\label{fig:Attention}
\end{figure}

\subsection{Visualization of the Color Queries}
We employ visualizations of the learned color queries to investigate their functionality. This visualization technique involves sigmoiding and scaling the dot product between an individual color query and the image feature map. As we can see in Fig. \ref{fig:Attention}, the results demonstrate that distinct color queries elicit unique regions of pronounced activation within the image feature map.

Taking the second line as an example, the first color query specifically attends to the diver's suit, capturing the black color embedding. The second and third color queries, on the other hand, focus on the background elements such as the reef and the fish, capturing the blue and silver-white color embeddings respectively. As another example, in the fourth row, the first color query directs its attention towards the spots on the manta ray's body, while the second and third color queries emphasize the white color of the manta ray's body and the sand that shares a similar hue, as well as the orange-red color present at the edges of the body.

\begin{table}[t]
	\huge
	\centering
	\caption{Quantitative ablation results of the proposed FDCE-Net on UIEB dataset.}
	\resizebox{\linewidth}{!}{
		\begin{tabular}{cccccc}
			\toprule[\heavyrulewidth]
			Model & W/o FS-Net & W/o FCE & W/o SCE & W/o DCE & Full Model \\
			\hline
			\hline
			\addlinespace[3pt]
			SSIM  & 0.911 & 0.908 & 0.913 & 0.904 & \textbf{0.917} \\
			PSNR  & 22.01 & 22.28 & 22.53 & 21.94 & \textbf{23.87} \\
			MSE   & 0.539 & 0.528 & 0.496 & 0.574 & \textbf{0.419} \\
			\bottomrule[\heavyrulewidth]
		\end{tabular}%
	}
	\label{tab:table3}%
	
\end{table}

\begin{table}[t]
	\centering
	\caption{{Performance in terms of PSNR and SSIM on LSUI dataset when using different feature fusion method in FSRB. The \textbf{best} results are shown in boldface.}}
	\begin{tabular}{ccccc}
		\toprule[1pt]
		Fusion & Addition & Attention & Bilinear & Concatenation \\
		\hline
		\hline
		\addlinespace[3pt]
		SSIM  & 0.817 & 0.877 & 0.803 & \textbf{0.881} \\
		PSNR  & 20.89 & 24.23 & 20.39 & \textbf{24.47} \\
		\bottomrule[1pt]
	\end{tabular}%
	\label{tab:FSRB}%
\end{table}
\begin{figure}[h]
	\centering
	\includegraphics[width=0.48\textwidth]{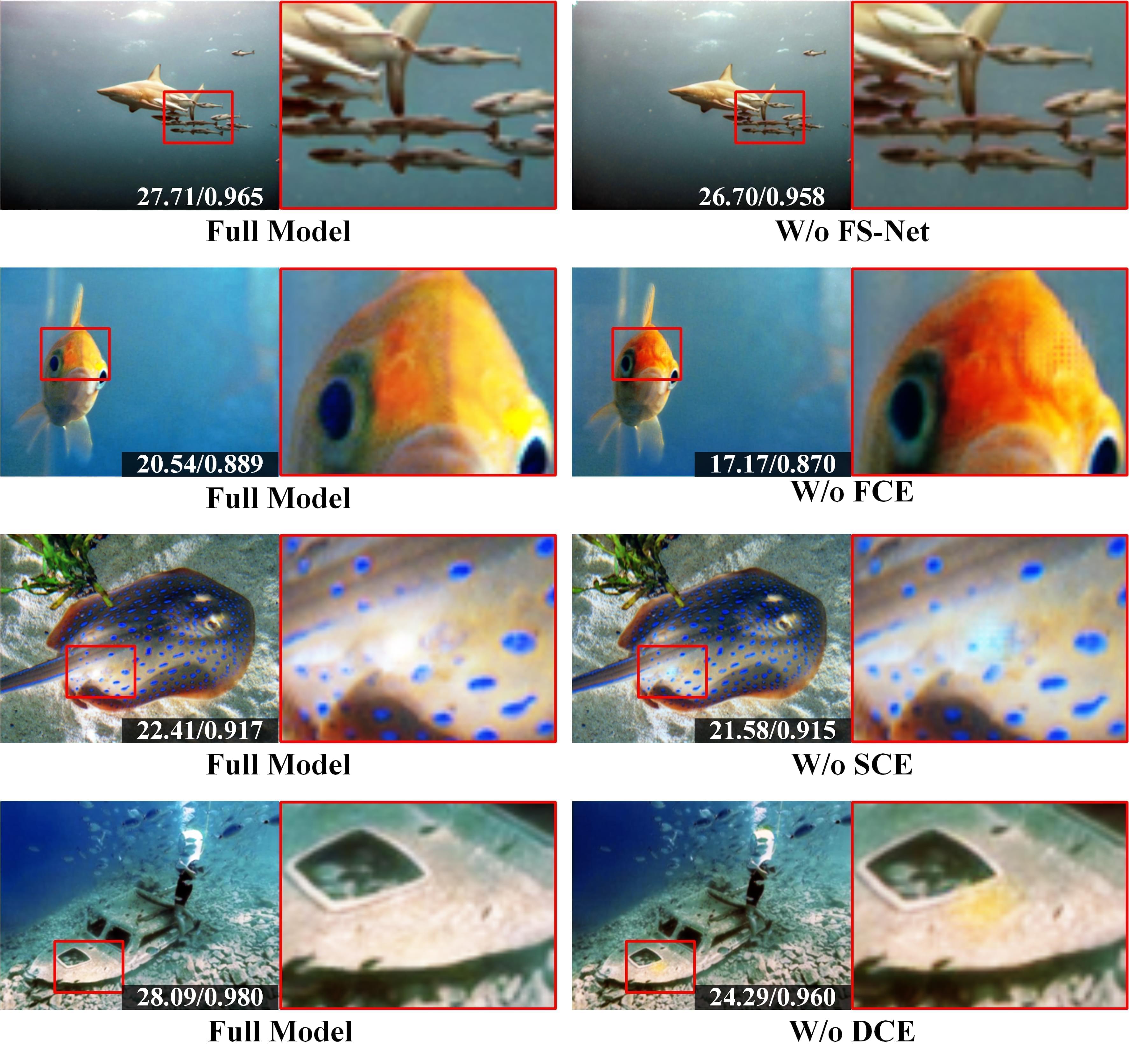}
	\caption{Qualitative ablation results of the proposed FDCE-Net, the area circled by the red rectangle has been enlarged for better visual comparison in each pair of images.}
	\label{fig:Ablation}
\end{figure}
\begin{table}[t]
	\centering
	\caption{{Fine tuning of $\alpha$ and $\beta$ values in our loss function on LSUI dataset in terms of SSIM and PSNR, the \textbf{best} results are shown in boldface.}}
	\resizebox{\linewidth}{!}{
		\begin{tabular}{lccccccc}
			\toprule[1pt]
			$\alpha$ in B+$\alpha\mathcal{L}_{hist}$ & 0.1   & 0.2   & 0.3   & 0.4   & 0.5   & 0.6   & 0.7\\
			\hline
			\hline
			\addlinespace[3pt]
			PSNR  & 23.18 & 23.49 & 23.76 & 23.91 & \textbf{23.94} & 23.90 & 23.90\\
			\midrule[1pt]
			$\beta$ in C+$\beta\mathcal{L}_{per}$  & 0.01  & 0.02  & 0.03  & 0.04  & 0.05  & 0.06  & 0.07\\
			\hline
			\hline
			\addlinespace[3pt]
			PSNR  & 24.03 & 23.98 & 24.21 & 24.37 & \textbf{24.47} & 24.46 & 24.41\\
			\bottomrule[1pt]
		\end{tabular}%
	}
	\label{tab:Loss2}%
\end{table}
\subsection{Ablation Study on Components of FDCE-Net}
We conducted an assessment of the impact of FS-Net and DCE on the performance of FDCE-Net. DCE can be decomposed into a first color encoder and a second color encoder.

To demonstrate the effectiveness of FS-Net, we performed the UIE task on the UIEB dataset both with and without it. Fig. \ref{fig:Ablation} illustrates the UIE results for different models, where ``W/o FS-Net" represents the model without FS-Net. We observed color enhancement inaccuracies in the images enhanced without FS-Net, as the overall color of the image appeared yellowish and the edge details of the fish were blurred. This verifies that FS-Net plays a crucial role in enhancing the overall color and texture details of the image, as described in Tab. \ref{tab:table3}. It is evident that the performance significantly drops in the absence of FS-Net, validating the rationale and efficacy of the proposed FS-Net in the overall UIE task.

\begin{table}[h]
	\centering
	\caption{{Ablation study of total loss function on LSUI dataset in terms of SSIM and PSNR, the \textbf{best} results are shown in boldface.}}
	\begin{tabular}{ccccc}
		\toprule[1pt]
		$\mathcal{L}_{total}$ & A     & B     & C     & D \\
		\hline
		\hline
		\addlinespace[3pt]
		SSIM  & 0.865 & 0.871 & 0.879 & \textbf{0.881}\\
		PSNR  & 23.06 & 23.45 & 23.94 & \textbf{24.47}\\
		\bottomrule[1pt]
	\end{tabular}%
	\label{tab:Loss1}%
\end{table}
To validate the role of DCE, we conducted three experiments in which we individually removed the first color encoder, then removed the second color encoder, and finally removed both color encoders (i.e., DCE). In Tab. \ref{tab:table3} and Fig. \ref{fig:Ablation}, ``W/o FCE", ``W/o SCE," and ``W/o DCE" represent the results of these three experiments, respectively. When the first color encoder was removed, we observed a large red patch on the fish's head in the image, and the overall color of the image appeared darker. This indicates that the color histogram calculated by the first color encoder plays an important role in the overall color recovery. Removing the second color encoder resulted in the generation of unreasonable color patches in the image due to the lack of semantic information. Moreover, after removing DCE entirely, the overall image quality deteriorated significantly, with lower contrast and less pronounced details. The overall color recovery of the image became much more inaccurate and exhibited unnatural color patches.

\begin{figure}[t]
	\centering
	\includegraphics[width=0.48\textwidth]{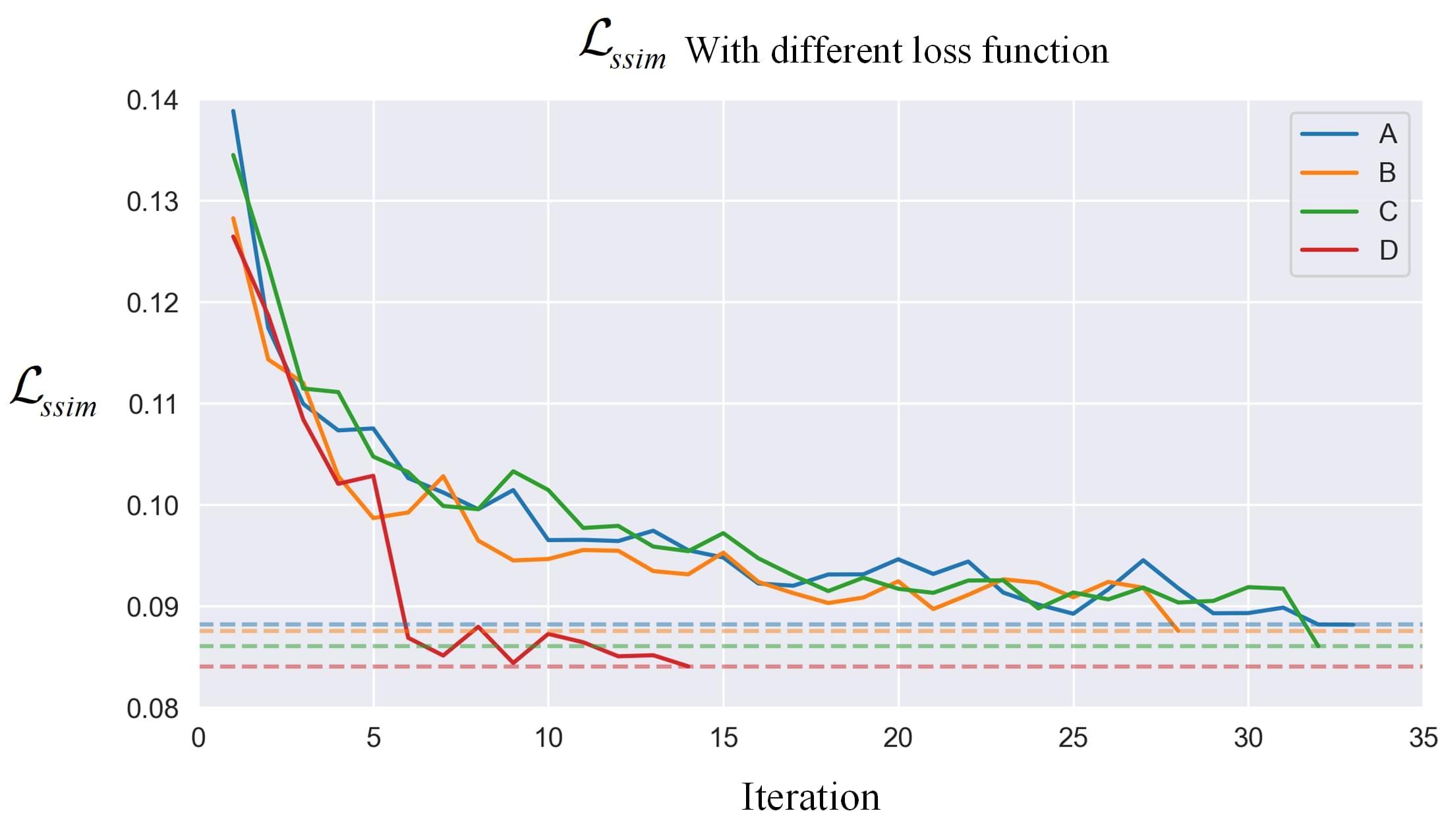}
	\caption{The speed of convergence and $\mathcal{L}_{ssim}$ obtained for different combinations of loss functions are compared by means of line plots.}
	\label{fig:Loss}
\end{figure}

\begin{table}[h]
	\centering
	\caption{{Experiments on the number of decoder groups N in the Second Color Decoder, N=3 is the result of balancing performance and computational complexity.}}
	\begin{tabular}{cccccc}
		\toprule[1pt]
		N     & 1     & 2     & 3     & 4     & 5\\
		\hline
		\hline
		\addlinespace[3pt]
		PSNR  & 23.91 & 24.18 & \textbf{24.47} & 24.40 & 24.41\\
		FLOPs & 12.51G & 12.64G & \textbf{12.78G} & 12.91G & 13.04G \\
		Params & 3.27M & 4.32M & \textbf{5.53M} & 6.43M & 7.48M\\
		\bottomrule[1pt]
	\end{tabular}%
	\label{tab:N}%
\end{table}
\begin{table*}[t]
	\centering
	\caption{{The detection accuracy of different underwater enhancement methods on the Aquarium dataset in terms of mAP (\%)↑ and mIoU (\%)↑. The \textbf{best} results are shown in boldface.}}
	\resizebox{\linewidth}{!}{
		\begin{tabular}{cccccccccccc}
			\toprule[1pt]
			& Objects & UWCNN & FUnIE-GAN & Ucolor & UHDSF-Net & PUIE\_MC & MMLE  & TACL  & Semi-UIR & WWPE  & Ours \\
			\hline
			\hline
			\addlinespace[3pt]
			\multirow{2}[2]{*}{YOLOv5} & mAP   & 70.48 & 71.69 & 73.86 & 80.72 & 79.93 & 79.52 & 76.86 & 75.34 & 79.05 & \textbf{84.56} \\
			& mIoU  & 72.11 & 74.43 & 77.61 & 83.47 & 81.58 & 80.47 & 78.64 & 76.48 & 81.26 & \textbf{86.89} \\
			\bottomrule[1pt]
		\end{tabular}%
	}
	\label{tab:Detection}%
\end{table*}
\begin{figure*}[t]
	\centering
	\includegraphics[width=1\linewidth]{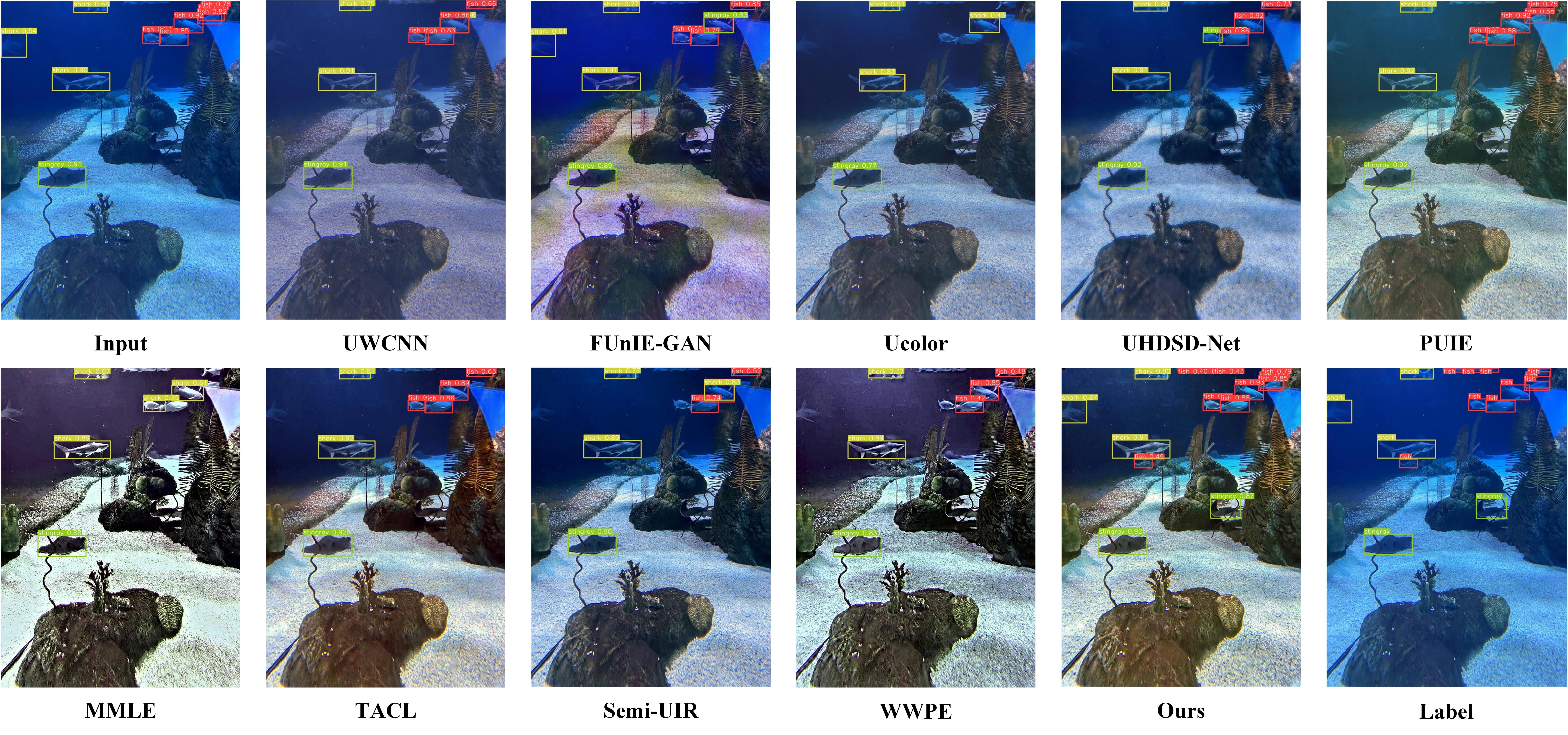}
	\captionsetup{font=footnotesize}
	\caption{Assessment of object detection on the Aquarium dataset using various enhancement techniques, utilizing the YOLOv5 detection algorithm. Notably, outcome enhanced by our method demonstrates considerable benefits in enhancing detection performance compared to alternative approaches.}
	\label{Detection}
\end{figure*}

Furthermore, our observations from Tab. \ref{tab:table3} confirm the importance of the first and second color encoders and their role in the overall performance. When the first and second color encoders were individually removed, we observed varying degrees of performance degradation in terms of average metrics. However, the removal of DCE resulted in a significant decline in average metrics, reaffirming the effectiveness of DCE. The first color encoder, along with the color histogram constraint, contributed effectively to the final results by capturing important color information. Similarly, the second color encoder played a crucial role in the color enhancement results by learning adaptive color queries from different semantic features.

\subsection{Ablation Study of Feature Aggregation Method in FS-Net}
{In the purpose of investigating different feature fusion methods’ impact on the result of our network, we have done comparative experiments on four commonly used feature fusion methods, eg. Addition, Concatenation, Attention-based feature aggregation and Bilinear feature fusion, \ref{tab:FSRB} shows the outcomes of FS-Net using different feature fusion methods. The performance significantly dropped when choosing Addition or Bilinear fusion. This drop can be attributed to the direct combination of features through addition, which may lead to the loss of some feature information. When information from two feature maps is merged into the same space, it has the potential to cancel out or confuse one another. Both Addition and Bilinear fusion mix these features to a certain degree, which could diminish the distinctiveness between features. Concatenation, on the other hand, maintains the independence of each feature map, enabling subsequent layers to freely learn how to optimally combine these features while preserving their distinctiveness. Although Concatenation and Attention-based aggregation exhibit similar performance levels, the Attention mechanism invariably introduces a greater computational burden. For the reasons mentioned above, we select concatenation as the method for feature fusion within our FSRB.}

\subsection{Ablation Study of Different combinations of loss function}
In order to verify the effectiveness of our loss function, we conducted experiments for different combinations of loss functions, A, B, C and D represent $\mathcal{L}_{ssim}$, $\mathcal{L}_{ssim}+\mathcal{L}_{rec}$, $\mathcal{L}_{ssim}+\mathcal{L}_{rec}+\alpha\mathcal{L}_{hist}$ and the full loss function we have chosen: $\mathcal{L}_{ssi m}+\mathcal{L}_{rec}+\alpha\mathcal{L}_{hist}+\beta\mathcal{L}_{per}$, respectively, 
from Fig. \ref{fig:Loss}, it can be observed that the loss function we have chosen achieves better metrics and requires less convergence time compared to the other combinations.

{For the fine-tuning of $\alpha$ and $\beta$, we set the loss functions as B+$\alpha\mathcal{L}_{hist}$ and C+$\beta\mathcal{L}_{per}$, respectively. As shown in Tab. \ref{tab:Loss2}, we set $\alpha$ to 0.5 and $\beta$ because they yielded the best experimental results. Similarly, according to the experimental results in Tab. \ref{tab:Loss1}, our total loss achieved the best experimental results compared to the combination of its subsets.}

\subsection{Ablation Study of Group Number N}
{In the second color encoder, we opted for an N value of 3 after weighing up performance and computational complexity. We experimented with various N values, ranging from 1 to 5. As demonstrated in Tab. \ref{tab:N}, performance peaked at N=3, with no further improvement observed as the number of groups increased. Therefore, we settled on an N value of 3 for our final model, as it delivers optimal performance without unnecessary redundancy.}

\section{Application}
{We evaluated the contribution of our FDCE-Net to underwater object detection. In this experiment, we selected 200 images containing objects from the UIEB dataset and the EUVP dataset. Firstly, we separately pre-trained object detection models using YOLOv5, and conducted experiments on object detection using the Aquarium\footnote{https://public.roboflow.com/object-detection/aquarium} dataset. We enhanced these images with our FDCE-Net and other SOTA UIE methods for comparison. Tab. \ref{tab:Detection} and Fig. \ref{Detection} show that the detection accuracy was superior after being processed by FDCEE-Net compared to other competing methods. Therefore, our model benefits both human vision and machine-based tasks.}

\section{Conclusion and Limitation}
The success of our approach can be attributed in part to the insights gained from analyzing the frequency domain characteristics of underwater images. Our FS-Net, with its unique design for handling color luminance and noise texture in the Fourier domain, plays a crucial role in our task. By leveraging the initial enhancement provided by FS-Net, we further introduce DCE to address the color aspect of UIE. The first color encoder learns color histograms and extracts multi-scale information from the images, while the second color encoder learns semantically-aware color embeddings through adaptive color queries. Through the collaborative efforts of these two color encoders, we achieve high-quality enhancement results for underwater images. Extensive experiments conducted on various benchmark datasets provide substantial evidence for the superiority of our solution. Ablation studies further validate the effectiveness of the key components in our approach.

{Similar to other supervised models, our model's training heavily depends on paired dataset construction. Additionally, excessive training may pose the risk of overfitting, potentially diminishing the generalization performance of pretrained models. As indicated in Tab. \ref{tab:SOTA_Comparison}, our model's superiority on unpaired datasets is not notably pronounced. However, exploring the integration of our approach with unsupervised, semi-supervised, and contrastive learning methodologies presents an intriguing direction for future research.}
	
	\bibliographystyle{IEEEtran}
	\bibliography{reference} 
	
	\begin{IEEEbiography}[{\includegraphics[width=1in,height=1.25in,clip,keepaspectratio]{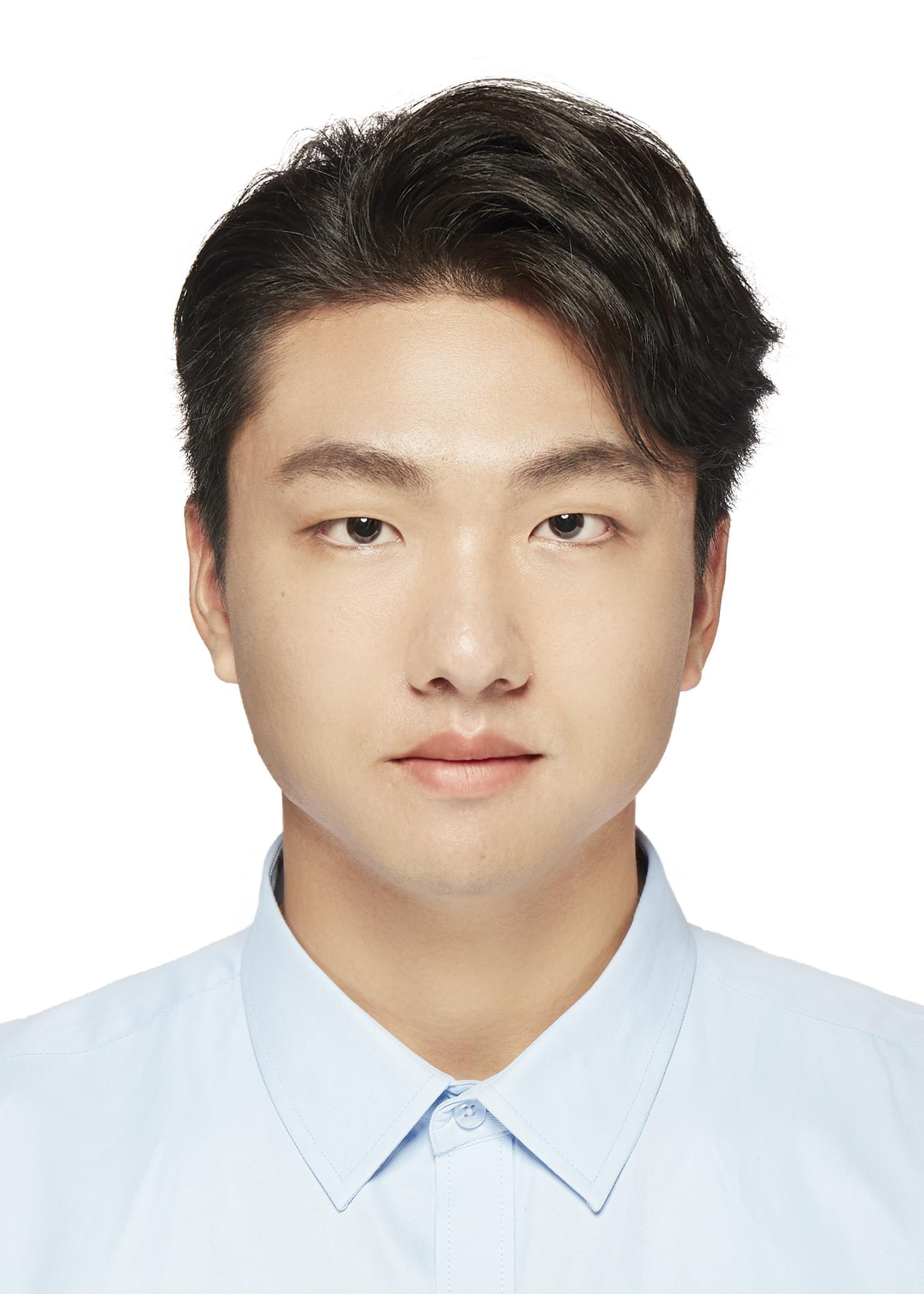}}]{Zheng Cheng}
		received the B.S. degree in Internet Engineering from Shandong University of Finance and Economics in 2019. Now he is pursuing the
		master’s degree with the College of Computer Science \& Technology, Qingdao University. His current research interests include computer vision and image processing.
	\end{IEEEbiography}

	\begin{IEEEbiography}[{\includegraphics[width=1in,height=1.25in,clip,keepaspectratio]{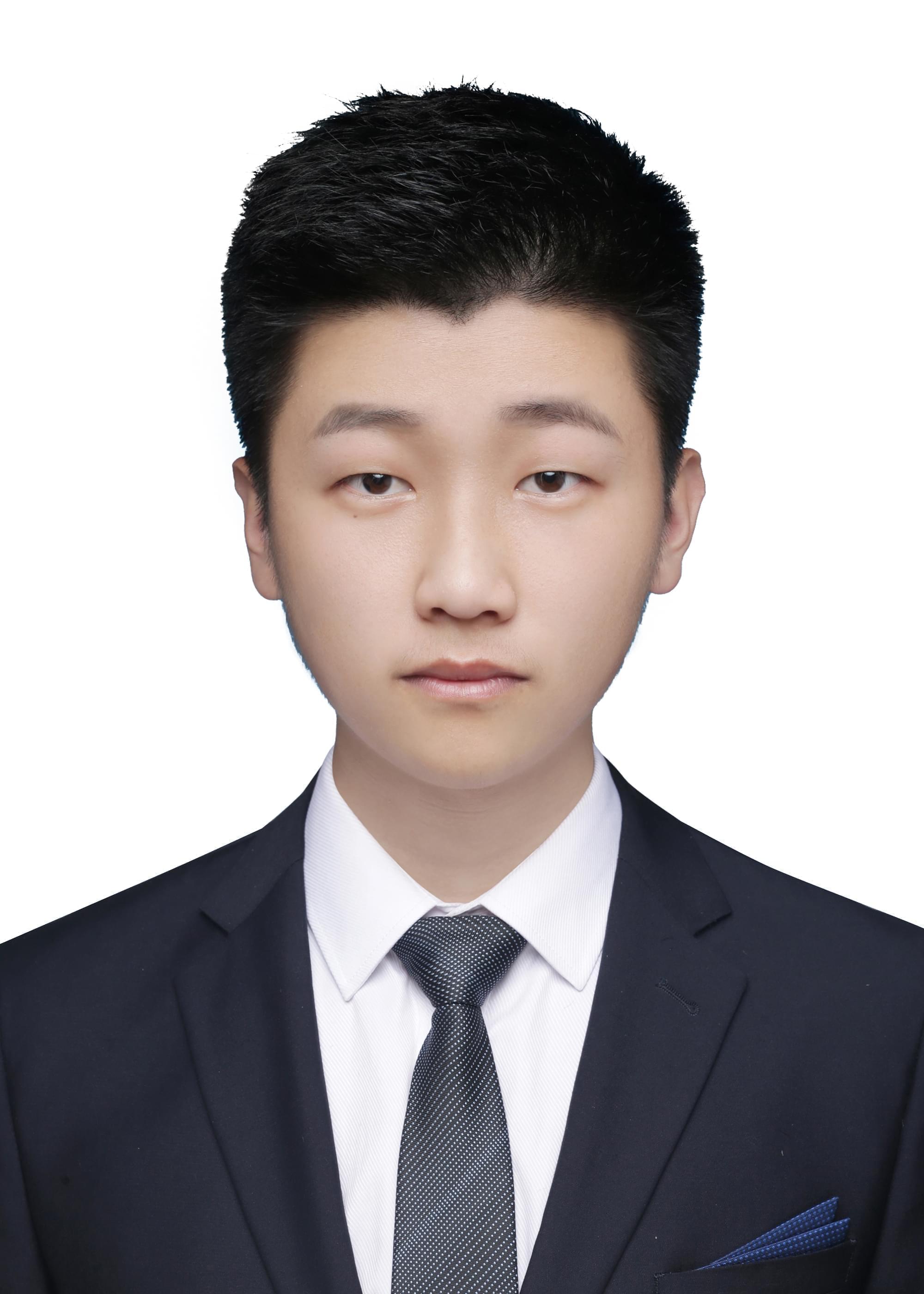}}]{Guodong Fan}
		received the M. Eng. degree from Shandong Technology and Business University, China, in 2021, where he is currently pursuing the Ph.D. degree with the Qingdao University.  His research interests are in image processing, machine learning, and computer vision.
	\end{IEEEbiography}

	\begin{IEEEbiography}[{\includegraphics[width=1in,height=1.25in,clip,keepaspectratio]{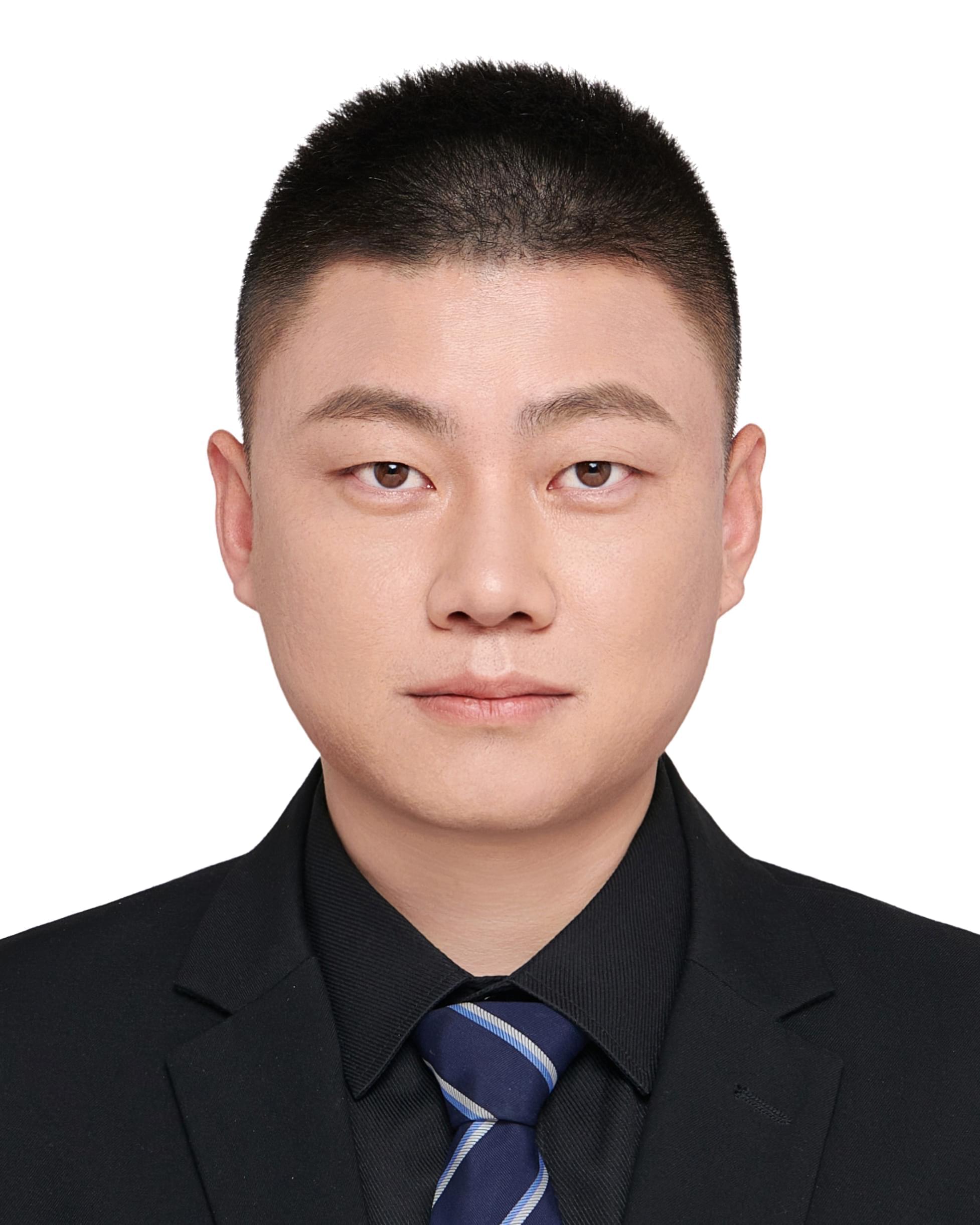}}]{Jingchun Zhou}
		received his M.S. degrees in Software Engineering from Beijing University of Posts and Telecommunications (BUPT) in 2016, the Ph.D. degree in Computer Applications from Dalian Maritime University (DMU), Dalian, China, in 2021. He is currently a Postdoctoral Researcher at DMU and a Visiting Scholar at the Hong Kong Polytechnic University (PolyU).
		
		He has authored over 40 research papers in esteemed journals and refereed conferences, such as the International Journal of Computer Vision (IJCV), IEEE Transactions on Geoscience and Remote Sensing (TGRS), Engineering Applications of Artificial Intelligence (EAAI), and more. Additionally, he has served as a reviewer for various prestigious journals, including IEEE Transactions on Image Processing, IEEE Transactions on Circuits and Systems for Video Technology, IEEE Transactions on Geoscience and Remote Sensing, IEEE Signal Processing Letters, IEEE Journal of Oceanic Engineering, Artificial Intelligence, Engineering Applications of Artificial Intelligence, Information Fusion, Applied Mathematical Modelling, Computers and Electronics in Agriculture, Signal Processing: Image Communication, Neurocomputing, Journal of Oceanic Engineering, Optics Express, among others.
		
		His research interests focus on computer vision, deep learning, and underwater image enhancement.
	\end{IEEEbiography}

	\begin{IEEEbiography}[{\includegraphics[width=1in,height=1.25in,clip,keepaspectratio]{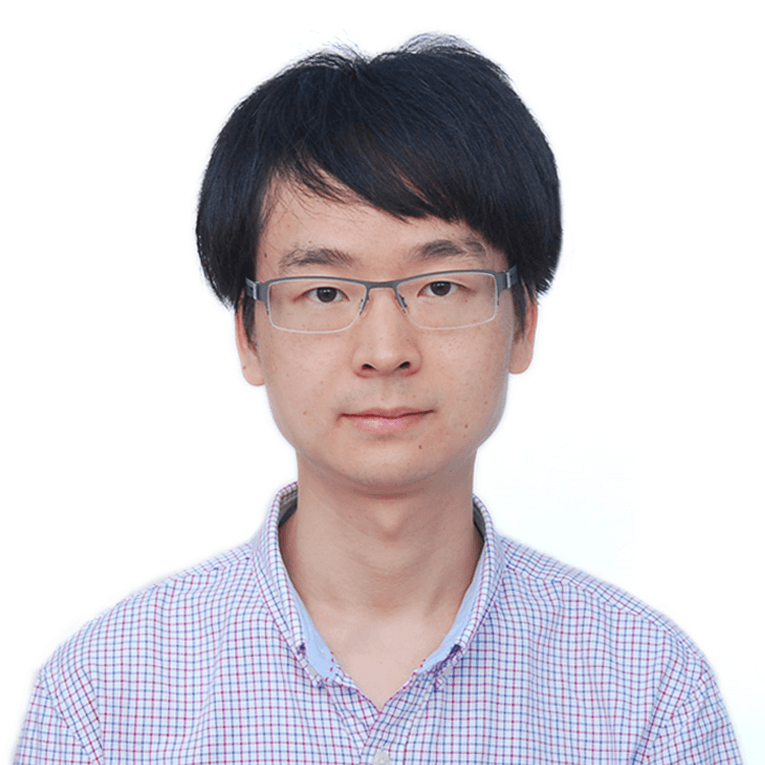}}]{Min Gan}
		received the B.S. degree in Computer Science and Engineering from Hubei University of Technology, Wuhan, China, in 2004, and the Ph.D. degree in Control Science and Engineering from Central South University, Changsha, China, in 2010.  His current research interests include machine learning, inverse problems, and image processing.
	\end{IEEEbiography} 
	
	\begin{IEEEbiography}[{\includegraphics[width=1in,height=1.25in,clip,keepaspectratio]{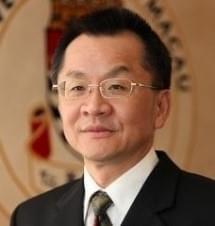}}]{C. L. Philip Chen}
		is the Chair Professor and Dean of the College of Computer Science and Engineering, South China University of Technology. Being a Program Evaluator of the Accreditation Board of Engineering and Technology Education(ABET) in the U.S., for computer engineering, electrical engineering, and software engineering programs, he successfully architects the University of Macau's Engineering and Computer Science programs receiving accreditations from Washington/Seoul Accord through Hong Kong Institute of Engineers(HKIE), of which is considered as his utmost contribution in engineering/computer Science education for Macau as the former Dean of the Faculty of Science and Technology. 
		Prof. Chen is a Fellow of IEEE, AAAS, IAPR, CAA, and HKIE; a member of Academia Europaea(AE), European Academy of Sciences and Arts(EASA), and International Academy of Systems and Cybernetics Science(IASCYS). He received IEEE Norbert Wiener Award in 2018 for his contribution in systems and cybernetics, and machine learnings.  He is also a Highly Cited Researcher by Clarivate Analytics in 2018 and 2021.
	\end{IEEEbiography}
	
\end{document}